\documentclass[letterpaper]{article}
\usepackage{aaai20}
\usepackage{times}
\usepackage{helvet}
\usepackage{courier}
\usepackage{soul}
\usepackage{url}
\usepackage[hidelinks]{hyperref}
\usepackage[utf8]{inputenc}
\usepackage[small]{caption}
\usepackage{graphicx}
\usepackage{amsmath}
\usepackage{booktabs}
\usepackage{algorithm}
\usepackage{algorithmic}
\usepackage[symbol]{footmisc}
\usepackage{epsfig}
\usepackage{array}
\usepackage{amssymb}
\usepackage{amsthm}
\usepackage{amsfonts}
\DeclareMathOperator*{\argmin}{arg\,min}
\DeclareMathOperator*{\argmax}{arg\,max}
\usepackage[position=top]{subfig}
\usepackage[export]{adjustbox}
\usepackage{wrapfig}
\usepackage{verbatim}
\newtheorem{theorem}{Proposition}[section]
\newtheorem{lemma}{Lemma}[section]
\usepackage{enumitem}
\mathchardef\mhyphen="2D
\usepackage{pifont}
\newcommand{\cmark}{\ding{51}}
\newcommand{\xmark}{\ding{55}}
\usepackage{color}
\newcommand{\citet}[1]{\citeauthor{#1} \shortcite{#1}}

\title{Robust Conditional GAN from Uncertainty-Aware Pairwise Comparisons}
\author{\Large \textbf{Ligong Han}\textsuperscript{\rm 1}, \Large \textbf{Ruijiang Gao}\textsuperscript{\rm 2}, \Large \textbf{Mun Kim}\textsuperscript{\rm 1}, \Large \textbf{Xin Tao}\textsuperscript{\rm 3}, \Large \textbf{Bo Liu}\textsuperscript{\rm 4}, \Large \textbf{Dimitris Metaxas}\textsuperscript{\rm 1}\\ 
\textsuperscript{\rm 1}Department of Computer Science, Rutgers University\\ \textsuperscript{\rm 2}McCombs School of Business, The University of Texas at Austin\\ \textsuperscript{\rm 3}Tencent YouTu Lab\quad \textsuperscript{\rm 4}JD Finance America Corporation\\ 
\texttt{l.han@rutgers.edu}\quad \texttt{ruijiang@utexas.edu}\quad \texttt{mun.kim@rutgers.edu}\\ \texttt{jiangsutx@gmail.com}\quad \texttt{kfliubo@gmail.com}\quad \texttt{dnm@cs.rutgers.edu} 
}

\begin{document}

\maketitle

\begin{abstract}
    Conditional generative adversarial networks have shown exceptional generation performance over the past few years. However, they require large numbers of annotations. To address this problem, we propose a novel generative adversarial network utilizing weak supervision in the form of pairwise comparisons (PC-GAN) for image attribute editing. In the light of Bayesian uncertainty estimation and noise-tolerant adversarial training, PC-GAN can estimate attribute rating efficiently and demonstrate robust performance in noise resistance. Through extensive experiments, we  show both qualitatively and quantitatively that PC-GAN performs comparably with fully-supervised methods and outperforms unsupervised baselines. Code can be found on the project website\footnote{\url{https://github.com/phymhan/pc-gan}}.
\end{abstract}

\section{Introduction}
Generative adversarial networks (GAN)~\cite{goodfellow2014generative} have shown great success in producing high-quality realistic imagery by training a set of networks to generate images of a target distribution via an adversarial setting between a generator and a discriminator. New architectures have also been developed for adversarial learning  such as conditional GAN (CGAN)~\cite{mirza2014conditional,odena2016conditional,han2018learning} which feeds a class or an attribute label for a model to learn to generate images conditioned on that label. The superior performance of CGAN makes it favorable for many problems in artificial intelligence (AI) such as image attribute editing.

However, this task faces a major challenge from the lack of massive labeled images with varying attributes. Many recent works attempt to alleviate such problems using semi-supervised or unsupervised conditional image synthesis~\cite{lucic2019high}. These methods mainly focus on conditioning the model on categorical pseudo-labels using self-supervised image feature clustering. However, attributes are often continuous-valued, for example, the stroke thickness of MNIST digits. In such cases, applying unsupervised clustering would be difficult since features are most likely to be grouped by salient attributes (like identities) rather than any other attributes of interest. In this work, to disentangle the target attribute from the rest, we focus on learning from weak supervisions in the form of pairwise comparisons.

\noindent \textbf{Pairwise comparisons.}
Collecting human preferences on pairs of alternatives, rather than evaluating absolute individual intensities, is intuitively appealing, and more importantly, supported by evidence from cognitive psychology~\cite{furnkranz2010preference}. As pointed out by~\citet{yanpassive}, we consider relative attribute annotation because they are (1) easier to obtain than total orders, (2) more accurate than absolute attribute intensities, and (3) more reliable in application like crowd-sourcing. For example, it would be hard for an annotator to accurately quantify the attractiveness of a person's look, but much easier to decide which one is preferred given two candidates. Moreover, attributes in images are often subjective. Different annotators have different criteria in their mind, which leads to noisy annotations~\cite{xu2019deep}.

Thus, instead of assigning an absolute attribute value to an image, we allow the model to learn to rank and assign a relative order between two images~\cite{yanpassive,furnkranz2010preference}. This method alleviates the aforementioned problem of lacking continuously valued annotations by learning to rank using pairwise comparisons.
\begin{figure}[t]
    \centering
    \includegraphics[scale=.6]{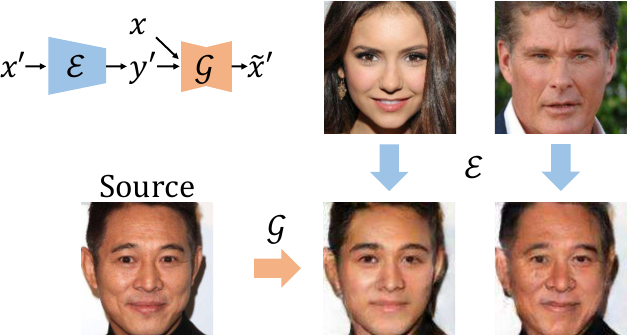}
    \caption{The generative process. Starting from a source image $x$, our model is able to synthesize a new image $\tilde{x}'$ with the desired attribute intensity possessed by the target image $x'$.}
    \label{fig:first}
\end{figure}

\noindent \textbf{Weakly supervised GANs.}
Our main idea is to substitute the full supervision with the attribute ratings learned from weak supervisions, as illustrated in Figure~\ref{fig:first}. To do so, we draw inspiration from the Elo rating system~\cite{elo1978rating} and design a Bayesian Siamese network to learn a rating function with uncertainty estimations. Then, for image synthesis, motivated by ~\cite{thekumparampil2018robustness} we use ``corrupted’’ labels  for adversarial training. The proposed framework can (1) learn from pairwise comparisons, (2) estimate the uncertainty of predicted attribute ratings, and (3) offer quantitative controls in the presence of a small portion of absolute annotations. Our contributions can be summarized as follows.
\begin{itemize}
    \item We propose a weakly supervised generative adversarial network, PC-GAN, from pairwise comparisons for image attribute manipulation. To the best of our knowledge, this is the first GAN framework considering relative attribute orders.
    \item We use a novel attribute rating network motivated from the Elo rating system, which models the latent score underlying each item and tracks the uncertainty of the predicted ratings.
    \item We extend the robust conditional GAN to continuous-value setting, and show that the performance can be boosted by incorporating the predicted uncertainties from the rating network.
    \item We analyze the sample complexity which shows that this weakly supervised approach can save annotation effort. Experimental results show that PC-GAN is competitive with fully-supervised models, while surpassing unsupervised methods by a large margin.
\end{itemize}

\section{Related Work}
\noindent \textbf{Learning to rank.}
Our work focuses on finding  ``scores'' for each item (e.g. player's rating) in addition to obtaining a ranking. The popular Bradley-Terry-Luce (BTL) model postulates a set of latent scores underlying all items, and the Elo system corresponds to the logistic variant of the BTL model. Numerous algorithms have been proposed since then. To name a few, TrueSkill~\cite{herbrich2007trueskill} considers a generalized Elo system in the Bayesian view. Rank Centrality~\cite{negahban2016rank} builds on spectral ranking and interprets the scores as the stationary probability under the random walk over comparison graphs. However, these methods are not designed for amortized inference, i.e. the model should be able to score (or extrapolate) an unseen item for which no comparisons are given. Apart from TrueSkill and Rank Centrality, the most relevant work is the RankNet~\cite{burges2005learning}. Despite being amortized, RankNet is homoscedastic and falls short of a principled justification as well as providing uncertainty estimations.

\noindent \textbf{Weakly supervised learning.}
Weakly-supervised learning focuses on learning from coarse annotations. It is useful because acquiring annotations can be very costly. A close weakly supervised setting to our problem is~\cite{xiao2015discovering} which learns the spatial extent of relative attributes using pairwise comparisons and gives an attribute intensity estimation. However, most facial attributes like attractiveness and age are not localized features thus cannot be exploited by local regions. In contrast, our work uses this relative attribute intensity for attribute transfer and manipulation.

\noindent \textbf{Uncertainty.}
There are two uncertainty measures one can model: aleatoric uncertainty and epistemic uncertainty. The epistemic uncertainty captures the variance of model predictions caused by lack of sufficient data; the aleatoric uncertainty represents the inherent noise underlying the data~\cite{kendall2017uncertainties}. In this work, we leverage Bayesian neural networks~\cite{gal2016dropout} as a powerful tool to model uncertainties in the Elo rating network.

\noindent \textbf{Robust conditional GAN (RCGAN).}
Conditioning on the estimated ratings, a normal conditional generative model can be vulnerable under bad estimations. To this end, recent research introduces noise robustness to GANs. \citet{bora2018ambientgan} apply a differentiable corruption to the output of the generator before feeding it into the discriminator. Similarly, RCGAN~\cite{thekumparampil2018robustness} proposes to corrupt the categorical label for conditional GANs and provides theoretical guarantees. Both methods have shown great denoising performance when noisy observations are present. To address our problem, we extend RCGAN to the continuous-value setting and incorporate uncertainties to guide the image generation.

\noindent \textbf{Image attribute editing.} There are many recent GAN-style architectures focusing on image attribute editing. IPCGAN~\cite{wang2018face} proposes an identity preserving loss for facial attribute editing. \citet{CycleGAN2017} propose cycle consistency loss that can learn the unpaired translation between image and attribute. BiGAN/ALI~\cite{donahue2016adversarial,dumoulin2016adversarially} learns an inverse mapping between image-and-attribute pairs.

There exists another line of research that is not GAN-based. Deep feature interpolation (DFI)~\cite{upchurch2017deep} relies on linear interpolation of deep convolutional features. It is also weakly-supervised in the sense that it requires two domains of images (e.g. young or old) with inexact annotations~\cite{zhou2017brief}. DFI demonstrates high-fidelity results on facial style transfer. While, the generated pixels look unnatural when the desired attribute intensity takes extreme values, we also find that DFI cannot control the attribute intensity quantitatively. \citet{wang2018weakly} considers a binary setting and sets qualitatively the intensity of the attribute. Unlike prior research, our method uses weak supervision in the form of pairwise comparisons and leverages uncertainty together with noise-tolerant adversarial learning to yield a robust performance in image attribute editing.

\section{Pairwise Comparison GAN}
In this section, we introduce the proposed method for pairwise weakly-supervised visual attribute editing. Denote an image collection as $I=\{x_1,\cdots, x_n\}$ and $x_i$'s underlying absolute attribute values as $\Omega\left(x_i\right)$. Given a set of pairwise comparisons $C$ (e.g., ${\Omega\left(x_i\right)>\Omega\left(x_j\right)}$ or ${\Omega\left(x_i\right)=\Omega\left(x_j\right)}$, where $i,j \in \{1,\cdots,n\}$), our goal is to generate a realistic image quantitatively with a different desired attribute intensity, for example, from 20 years old to 50 years old. The proposed framework consists of an Elo rating network followed by a noise-robust conditional GAN.

\subsection{Attribute Rating Network}
The designed attribute rating module is motivated by the Elo rating system~\cite{elo1978rating}, which is widely used to evaluate the relative levels of skills between players in zero-sum games. Elo rating from a player is represented as a scalar value which is adjusted based on the outcome of games. We apply this idea to image attribute editing by considering each image as a player and comparison pairs as games with outcomes. Then we learn a rating function.

\begin{figure}[t]
    \centering
    \includegraphics[scale=.4]{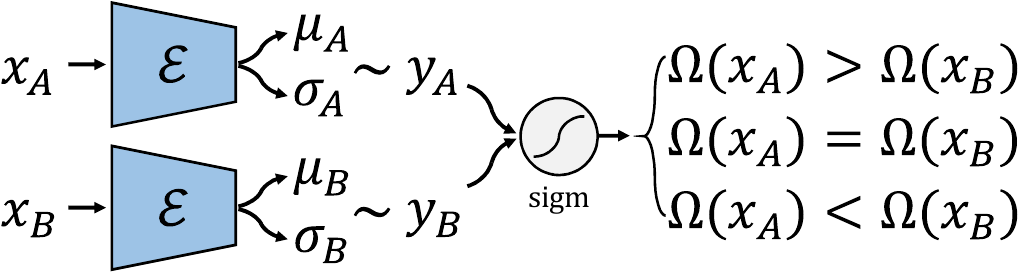}
    \caption{The Elo rating network. The comparison is performed by feeding into a sigmoid function the difference of ratings (scalar) of a given image pair. After training, the encoder $\mathcal{E}$ is used to train the PC-GAN, as illustrated in Figure~\ref{fig:gan_overview}.}
    \label{fig:elo_arch}
\end{figure}

\noindent \textbf{Elo rating system.} The Elo system assumes the performance of each player is normally distributed. For example, if \texttt{Player A} has a rating of $y_A$ and \texttt{Player B} has a rating of $y_B$, the probability of \texttt{Player A} winning the game against \texttt{Player B} can be predicted by $P_A = \frac{1}{1+10^{(y_B-y_A)/400}}$. We use $S_A$ to denote the actual score that \texttt{Player A} obtains after the game, which can be valued as $S_A(\texttt{win})=1, S_A(\texttt{tie})=0.5, S_A(\texttt{lose})=0.$
After each game, the player's rating is updated according to the  difference between the prediction $P_A$ and the actual score $S_A$ by $y_A'=y_A+K(S_A-P_A)$, where $K$ is a constant.

\noindent \textbf{Image pair rating prediction network.} Given an image pair $(x_A, x_B)$ and a certain attribute $\Omega$, we propose to use a neural network for predicting the relative attribute relationship between $\Omega(x_A)$ and $\Omega(x_B)$. This design allows amortized inference, that is, the rating prediction network can provide ratings for both seen and unseen data. The model structure is illustrated in Figure~\ref{fig:elo_arch}.

The network contains two input branches fed with $x_A$ and $x_B$. For each image $x$, we propose to learn its rating value $y_x$ by an encoder network $\mathcal{E}(x)$. Assuming the rating value of $x$ follows a normal distribution, that is $y_x\sim \mathcal{N}\left(\mu(x),\sigma^2(x)\right)$, we employ the reparameterization trick~\cite{kingma2013auto}, $y_x=\mu(x)+\epsilon \sigma(x)$ (where $\epsilon\sim\mathcal{N}(0,I)$). After obtaining each image's latent rating $y_A$ and $y_B$, we formulate the pair-wise attribute comparison prediction as $P_{A,y}(\Omega(x_A)>\Omega(x_B)|x_A,x_B,y_A,y_B) = \text{sigm}(y_A-y_B)$ where $\text{sigm}$ is the sigmoid function. Then, the predictive probability of $x_A$ winning $x_B$ is obtained by integrating out the latent variables $y_A$ and $y_B$,
\begin{align}
    P_A(\Omega(x_A)>\Omega(x_B)|x_A,x_B) = \int{\text{sigm}(y_A-y_B) dy_A dy_B},
    \label{eq:elo_expect}
\end{align}
\noindent and $P_B=1-P_A$. The above integration is intractable, and can be approximated by Monte Carlo, $P_A \approx P_A^{MC} = \frac{1}{M}\sum_{m=1}^{M}{P_{A,m}}$. We denote the ground-truth of $P_A$ and $P_B$ as $S_A$ and $S_B$. The ranking loss $\mathcal{L}_{rank}$ can be formulated with a logistic-type function, that is
\begin{align}
    \mathcal{L}_{rank}^{MC} = -\mathbb{E}_{x_A,x_B\sim C}{\left[S_A \log P_A^{MC} + S_B \log P_B^{MC}\right]}.
\end{align}
\noindent Noticing that $\mathcal{L}_{rank}^{MC}$ is biased, an alternative unbiased upper bound can be derived as
\begin{align}
    \mathcal{L}_{rank}^{UB} = -\mathbb{E}_{x_A,x_B\sim C}{\left[\frac{1}{M}\sum_{m=1}^{M}{S_A \log P_{A,m} + S_B \log P_{B,m}}\right]}.
\end{align}
\noindent In practice, we find that $\mathcal{L}_{rank}^{UB}$ performs slightly better than $\mathcal{L}_{rank}^{MC}$.

We further consider a Bayesian variant of $\mathcal{E}$. The Bayesian neural network is shown to be able to provide the epistemic uncertainty of the model by estimating the posterior over network weights in network parameter training~\cite{kendall2017uncertainties} . Specifically, let $q_{\theta}(w)$ be an approximation of the true posterior $p(w|\text{data})$ where $\theta$ denotes the parameter of $q$, we measure the difference between $q_{\theta}(w)$ and $p(w|\text{data})$ with the KL-divergence. The overall learning objective is the negative evidence lower bound (ELBO)~\cite{kingma2013auto,gal2016dropout},
\begin{align}
    \mathcal{L}_{\mathcal{E}} = \mathcal{L}_{rank}+\underbrace{\text{D}_{KL}(q_\theta(w) \Vert p(w|\text{data}))}_{{KL}}.
    \label{eq:elbo}
\end{align}
\citet{gal2016dropout} propose to view dropout together with weight decay as a Bayesian approximation, where sampling from $q_\theta$ is equivalent to performing dropout and the KL term in Equation~\ref{eq:elbo} becomes $L_2$ regularization (or weight decay) on $\theta$.

The predictive uncertainty of rating $y$ for image $x$ can be approximated using:
\begin{align}
    \hat{\sigma}^2(y) \approx \frac{1}{T}\sum_{t=1}^{T}{\mu_t^2}-(\frac{1}{T}\sum_{t=1}^{T}{\mu_t})^2+\frac{1}{T}\sum_{t=1}^{T}{\sigma_t^2}
\end{align}
\noindent with $\{\mu_t, \sigma_t\}_{t=1}^T$ a set of $T$ sampled outputs: $\mu_t, \sigma_t=\mathcal{E}(x)$.

\noindent \textbf{Transitivity.} Notice that the transitivity does not hold because of the stochasticity in $y$. If we fix $\sigma(\cdot)$ to be zero and a non-Bayesian version is used, the Elo rating network becomes a RankNet~\cite{burges2005learning}, and transitivity holds. However, one can still maintain transitivity by avoiding reparameterization and modeling $P_{A} = \text{sigm}(\frac{\mu(x_A)-\mu(x_B)}{\sqrt{\sigma^2(x_A)+\sigma^2(x_B)}})$. In practice, we find that reparameterization works better.

\begin{figure}[t]
    \centering
    \includegraphics[scale=.76]{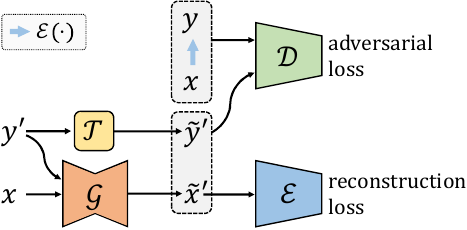}
    \caption{Overview of PC-GAN. Image $\tilde{x}'$ is synthesized from $x$ and $y'$. $y'$ is then ``corrupted'' to $\tilde{y}'$ by the transition $\mathcal{T}$, where $\mathcal{T}$ is a sampling process $\tilde{y}'\sim\mathcal{N}(y', \hat{\sigma}'^2)$. The reconstruction on attribute rating enforces mutual information maximization. The main difference between PC-GAN and a normal conditional GAN is that the conditioned rating of the generated sample is corrupted before feeding into the adversarial discriminator, forcing the generator to produce samples with clean ratings.}
    \label{fig:gan_overview}
\end{figure}

\subsection{Conditional GAN with Noisy Information}
We construct a CGAN-based framework for image synthesis conditioned on the learned attribute rating. The overall training procedure is shown in Figure~\ref{fig:gan_overview}: given a pair of images $x$ and $x'$, the generator $\mathcal{G}$ is trained to transform $x$ into $\tilde{x}'=\mathcal{G}(x,y')$, such that $\tilde{x}'$ possesses the same rating $y'=\mathcal{E}(x')$ as $x'$. The predicted ratings can still be noisy, thus a robust conditional GAN is considered. While RCGAN~\cite{thekumparampil2018robustness} is conditioned on discrete categorical labels that are ``corrupted’’ by a confusion matrix, our model relies on the ratings that are continuous-valued and realizes the ``corruption’’ via resampling.

\noindent \textbf{Adversarial loss.} Given image $x$, the corresponding rating $y$ can be obtained from a forward pass of the pre-trained encoder $\mathcal{E}$. Thus $\mathcal{E}$ defines a joint distribution $p_{\mathcal{E}}(x,y)=p_{data}(x)p_{\mathcal{E}}(y|x)$. Importantly, the output $\tilde{x}'$ of $\mathcal{G}$ is paired with a corrupted rating $\tilde{y}'=\mathcal{T}(y')$, where $\mathcal{T}$ is a sampling process $\tilde{y}'\sim\mathcal{N}(y', \hat{\sigma}'^2)$. The adversarial loss is
\begin{align}
    \mathcal{L}_{CGAN} = &\mathbb{E}_{x,y\sim p(x,y)}{\log(\mathcal{D}(x,y))} \quad+ \\
    &\mathbb{E}_{x\sim p(x), y'\sim p(y'), \tilde{y}'\sim \mathcal{T}(y')}{\log(1-\mathcal{D}(\mathcal{G}(x,y'),\tilde{y}'))}. \nonumber
\end{align}

The discriminator $\mathcal{D}$ is discriminating between real data $(x, y)$ and generated data $(\mathcal{G}(x,y'),\mathcal{T}(y'))$. At the same time, $\mathcal{G}$ is trained to fool $\mathcal{D}$ by producing images that are both realistic and consistent with the given attribute rating. As such, the Bayesian variant of the encoder is required for considering robust conditional adversarial training.

\noindent \textbf{Mutual information maximization.} Besides conditioning the discriminator, to further encourage the generative process to be consistent with ratings and thus learn a disentangled representation~\cite{chen2016infogan}, we add a reconstruction loss on the predictive ratings:
\begin{align}
    \mathcal{L}_{rec}^{y} = \mathbb{E}_{x\sim p(x), y'\sim p(y')}{\frac{1}{2\hat{\sigma}'^2}\Vert \mathcal{E}(\mathcal{G}(x,y'))-y'\Vert^2_2+\frac{1}{2}\log{\hat{\sigma}'^2}}.
\end{align}
The above reconstruction loss can be viewed as the conditional entropy between $y'$ and $\mathcal{G}(x,y')$,
\begin{align}
    \mathcal{L}_{rec}^{y} &\propto -\mathbb{E}_{y'\sim p(y'), \tilde{x}'\sim \mathcal{G}(x,y')}{\left[\log{p(y'|\tilde{x}')}\right]} \nonumber \\
    &= -\mathbb{E}_{y'\sim p(y'), \tilde{x}'\sim \mathcal{G}(x,y')}{\left[\mathbb{E}_{y\sim (y|\tilde{x}')}{\left[\log{(y|\tilde{x}')}\right]}\right]} \nonumber \\
    &= \text{H}(y'|\mathcal{G}(x,y')).
\end{align}
Thus, minimizing the reconstruction loss is equivalent to maximizing the mutual information between the conditioned rating and the output image.
\begin{align}
    \argmin_{\mathcal{G}}{\mathcal{L}_{rec}^{y}} &= \argmax_{\mathcal{G}}{-\text{H}(y'|\mathcal{G}(x,y'))} \nonumber \\
    &= \argmax_{\mathcal{G}}{-\text{H}(y'|\mathcal{G}(x,y'))+\text{H}(y')} \nonumber \\
    &= \argmax_{\mathcal{G}}{\text{I}(y';\mathcal{G}(x,y'))}.
\end{align}

The cycle consistency constraint forces the image $\mathcal{G}(\tilde{x}',y)$ to be close to the original $x$, and therefore helps preserve the identity information. Following the same logic, the cycle loss can be also viewed as maximizing the mutual information between $x$ and $\mathcal{G}(x,y')$.

\noindent \textbf{Full objective.} Finally, the full objective can be written as:
\begin{align}
    \mathcal{L}(\mathcal{G}, \mathcal{D}) = \mathcal{L}_{CGAN} + \lambda_{rec}\mathcal{L}_{rec}^{y} + \lambda_{cyc}\mathcal{L}_{cyc},
\end{align}
where $\lambda$s control the relative importance of corresponding losses. The final objective formulates a minimax problem where we aim to solve:
\begin{align}
    \mathcal{G}^* = \arg\min_{\mathcal{G}}{\max_{\mathcal{D}}{\mathcal{L}(\mathcal{G}, \mathcal{D})}}.
\end{align}

\noindent \textbf{Analysis of loss functions.} \citet{goodfellow2014generative} show that the adversarial training results in minimizing the Jensen-Shannon divergence between the true conditional and the generated conditional. Here, the approximated conditional will converge to the distribution characterized by the encoder $\mathcal{E}$. If $\mathcal{E}$ is optimal, the approximated conditional will converge to the true conditional, we defer the proof in Supplementary.

\noindent \textbf{GAN training.}
In practice, we find that the conditional generative model trains better if equal-pairs (pairs with approximately equal attribute intensities) are filtered out and only different-pairs (pairs with clearly different intensities) are remained. Comparisons of training CGAN with or without equal-pairs can be found in Supplementary.
\begin{table}[h]
    \centering
    \scalebox{0.8}{
    \begin{tabular}{lc|ccc}
    \hline
    Strategy           & {\bf Corr} &  {\bf IS} &        {\bf FID} & {\bf Acc (\%)} \\
    \hline
    rand+diff        & 0.91 & 3.65 $\pm$ 0.05 & 24.10 $\pm$ 0.24 &          67.44 \\
    rand+all         & 0.95 & 3.52 $\pm$ 0.03 & 21.75 $\pm$ 1.34 &          58.10 \\
    easy+diff        & 0.79 & 2.97 $\pm$ 0.05 & 29.55 $\pm$ 1.00 &          46.48 \\
    easy+all         & 0.81 & 2.82 $\pm$ 0.03 & 63.86 $\pm$ 1.32 &          51.46 \\
    hard+diff        & 0.92 & 2.90 $\pm$ 0.03 & 29.24 $\pm$ 1.07 &          43.78 \\
    hard+all         & 0.95 & 3.01 $\pm$ 0.04 & 22.04 $\pm$ 1.05 &          32.22 \\
    hard+pseudo-diff & 0.92 & 3.56 $\pm$ 0.02 & 26.03 $\pm$ 0.39 &          68.02 \\
    hard+pseudo-all  & 0.95 & 3.29 $\pm$ 0.03 & 24.94 $\pm$ 1.17 &          51.96 \\
    \hline
    \end{tabular}}
    \caption{Pair sampling strategies. Spearman correlations ({\bf Corr}), Inception Scores ({\bf IS}), Fr\'{e}chet Inception Distances ({\bf FID}), and classification accuracies ({\bf Acc}) evaluated on UTKFace are reported. \texttt{hard+diff} stands for training Elo rating with hard examples and training CGAN with different-pairs only, and \texttt{pseudo-diff} stands for the pairs augmented with pseudo-pairs but with equal pairs filtered out. If the same active learning strategy is used (e.g. \texttt{rand+diff} and \texttt{rand+all}), CGANs are conditioned on the same ratings trained from all pairs (e.g. \texttt{rand+all}).}
    \label{tab:online}
\end{table}

\subsection{Pair Sampling}
Active learning strategies such as OHEM~\cite{shrivastava2016training} can be incorporated in our Elo rating network. In hard example mining, only pairs with small rating differences are queried (\texttt{hard+diff/all} in Table~\ref{tab:online}). In addition, to maximize the number of different-pairs we also try easy example mining (\texttt{easy+diff/all} in Table~\ref{tab:online}). As shown, easy examples are inferior to hard examples in terms of both rating correlations and image qualities. The reason might be that easy example mining chooses pairs with drastic differences in attribute intensity, which makes the model hard to train. Hard examples help to learn a better rating function, however, provide less amount of different-pairs for the generative model to capture attribute transitions. We therefore augment hard examples with pseudo-pairs (easy examples but with predicted labels, listed as \texttt{hard+pseudo-diff/all} in Table~\ref{tab:online}). The augmentation strategy works well, but in following experiments we use randomly sampled pairs because (1) the random strategy is simple and performs equally well, and (2) pseudo-labels are less reliable than queried labels.

\noindent \textbf{Number of pairs.} Suppose there are $n$ images in the dataset, then the possible number of pairs is upper bounded by $n(n-1)/2$. However, if $\mathcal{O}(n^2)$ pairs are necessary, there is no benefit of choosing pairwise comparisons over absolute label annotation. Using results from \cite{radinsky2011ranking,wauthier2013efficient}, the following proposition shows that only $\mathcal{O}(n)$ comparisons are needed to recover an approximate ranking.
\begin{theorem}
For a constant $d$ and any $0<\lambda<1$, if we measure $dn/\lambda^2$ comparisons chosen uniformly with repetition, the Elo rating network will output a permutation $\hat{\pi}$ of expected risk at most $(2/\lambda)(n(n-1)/2)$.
\end{theorem}
We also provide an empirical study in the Supplementary that supports the above proposition.

\begin{figure}[h]
  \begin{center}
    \subfloat[t-SNE
    ]{\includegraphics[width=0.3\linewidth]{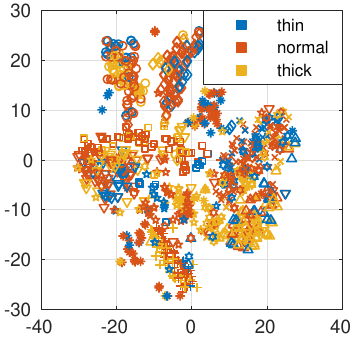}}
    \hspace{5pt}
    \subfloat[Ratings
    ]{\includegraphics[width=0.3\linewidth]{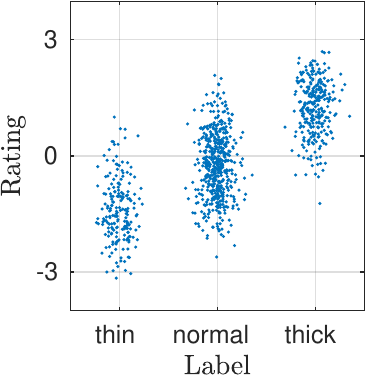}}
    \hspace{5pt}
    \subfloat[Samples
    ]{\includegraphics[width=0.28\linewidth]{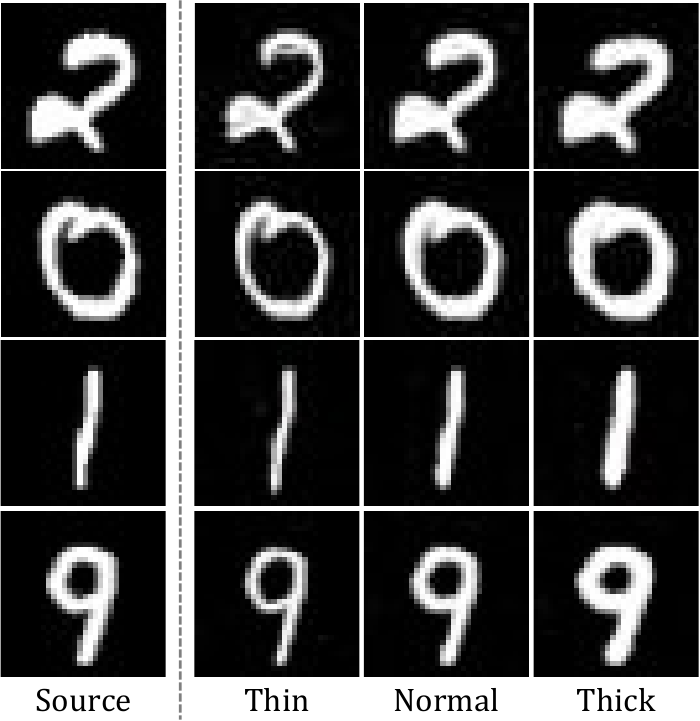}}
    \caption{Results on Annotated MNIST. (a) t-SNE visualization of the MNIST dataset, different shapes correspond to different numbers, and different colors represent various thickness levels. As shown, data is clustered by numbers rather than thickness. (b) Visualization of ratings learned from pairwise comparisons (ground-truth labels are jittered for better visualization). (c) Samples of thickness editing results.}
    \label{fig:mnist}
  \end{center}
\end{figure}
\begin{figure}[h]
    \centering
    \includegraphics[scale=0.48]{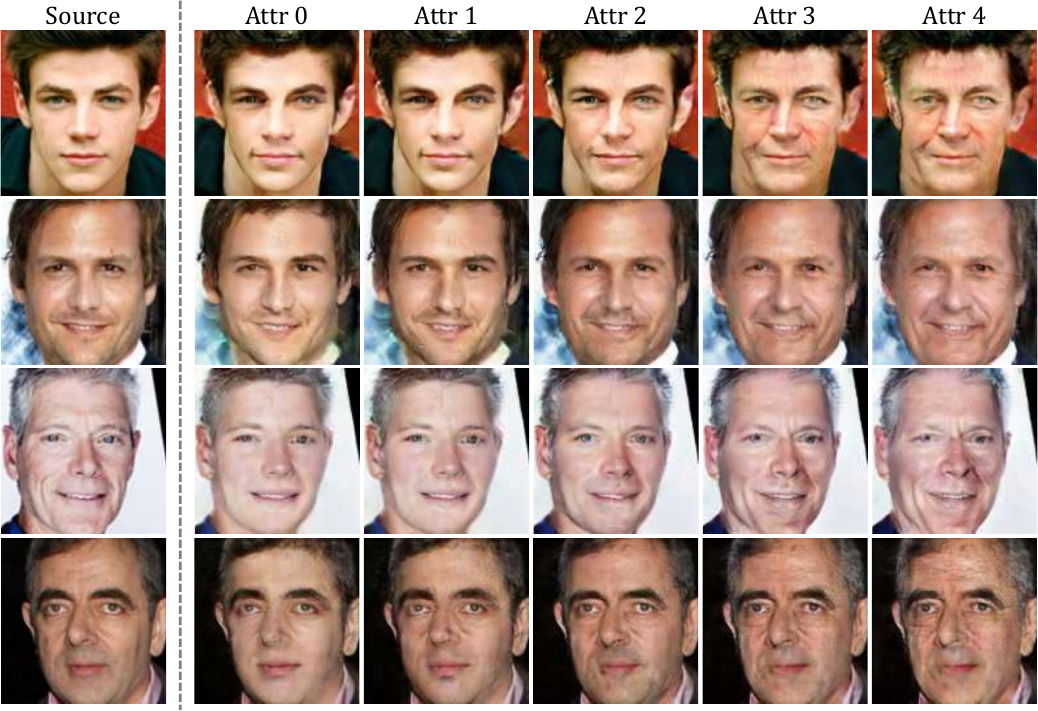}
    \caption{Results on CACD. The target attribute is age. Values from \texttt{Attr0} to \texttt{Attr4} correspond to age of $15$, $25$, $35$, $45$ and $55$, respectively.}
    \label{fig:results_cacd}
\end{figure}
\begin{figure}[h]
    \centering
    \includegraphics[scale=.48]{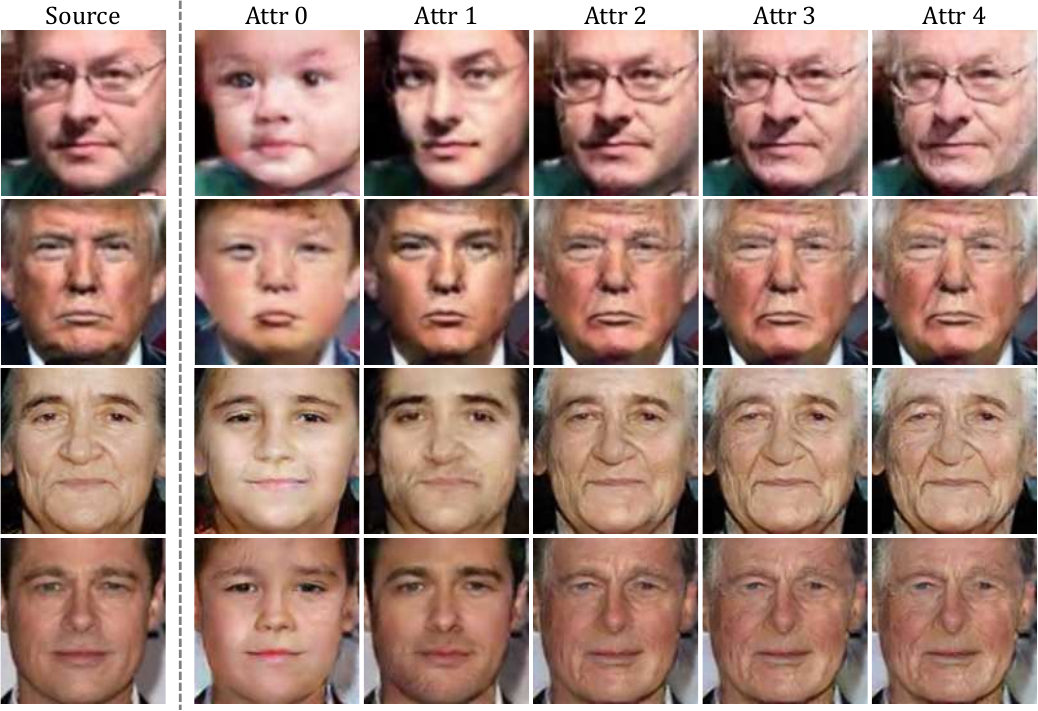}
    \caption{Results on UTKFace. The target attribute is age. Values from \texttt{Attr0} to \texttt{Attr4} correspond to age of $10$, $30$, $50$, $70$ and $90$, respectively.}
    \label{fig:results_utkface}
\end{figure}
\begin{figure}[h]
    \centering
    \includegraphics[scale=.48]{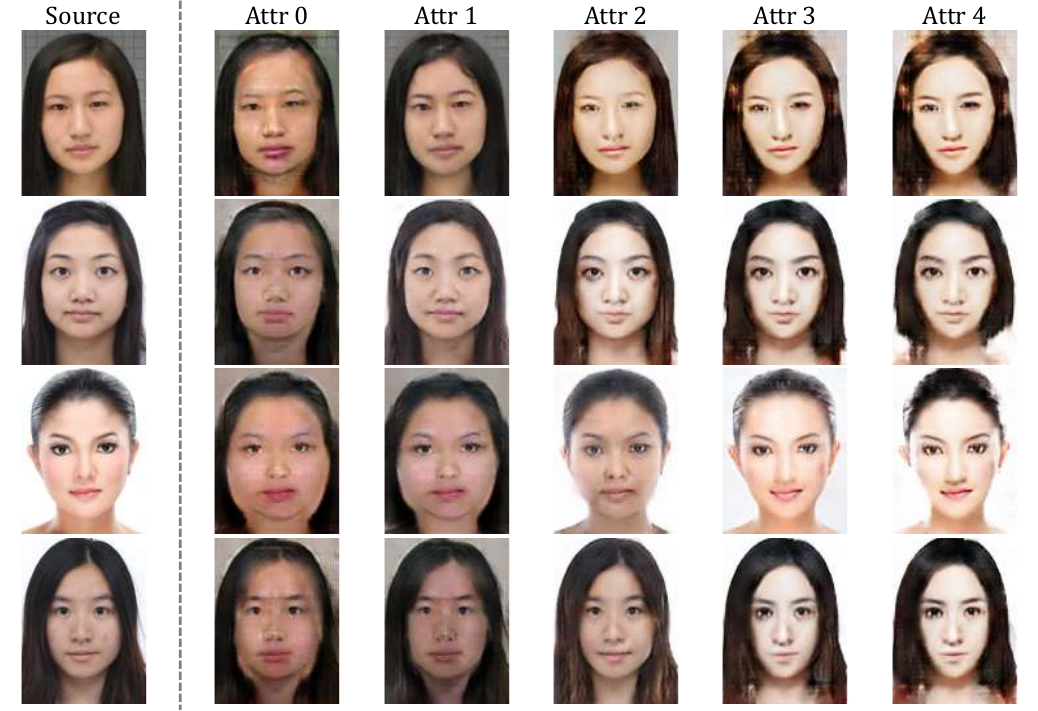}
    \caption{Results on SCUT-FBP. The target attribute is attractiveness score (1 to 5). Values from \texttt{Attr0} to \texttt{Attr4} correspond to score of $1.375$, $2.125$, $2.875$, $3.625$ and $4.5$, respectively.}
    \label{fig:results_scut}
\end{figure}
\begin{figure}[h]
    \centering
    \includegraphics[scale=.48]{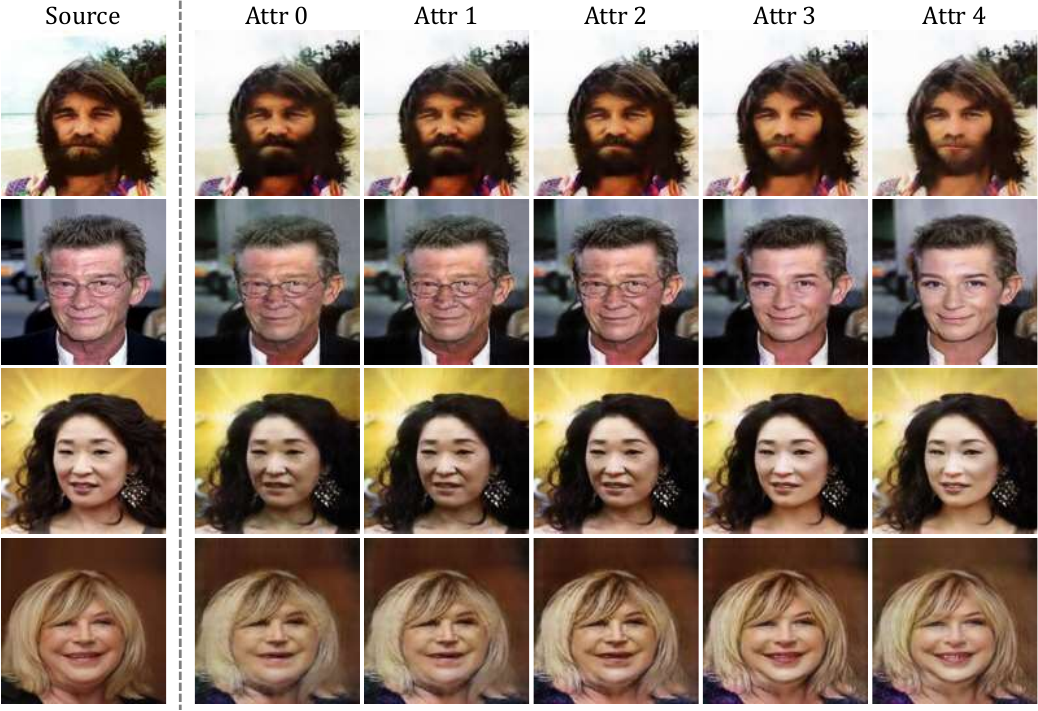}
    \caption{Results on CelebA. The target attribute is attractiveness. We take the cluster mean of ratings for ``attractive'' being -1 and 1 as \texttt{Attr0} and \texttt{Attr4} respectively. \texttt{Attr1} to \texttt{Attr3} are then linearly sampled. Results show a smooth transition of visual features, for example, facial hair, aging related features, smile lines, and shape of eyes.}
    \label{fig:results_celeba}
\end{figure}
\begin{figure*}[t]
    \centering
    \includegraphics[scale=.48]{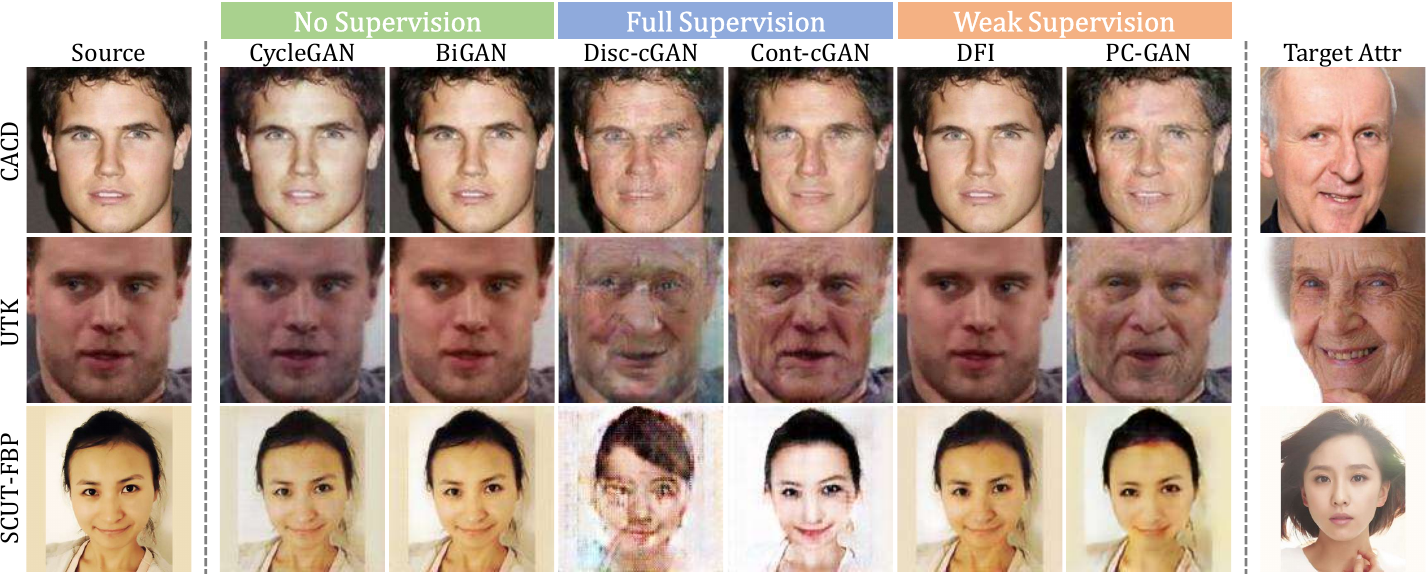}
    \caption{Baselines: (\textbf{left}) Source images from different datasets; (\textbf{right}) target images with desired attribute intensity; (\textbf{middle}) synthesized images by different methods to the desired attribute intensity. Unsupervised baselines cannot effectively change the attribute to the desired intensity.}
    \label{fig:baselines}
\end{figure*}
\section{Experiments}
In this section, we first present a motivating experiment on MNIST. Then we evaluate the PC-GAN in two parts: (1) learning attribute ratings, and (2) conditional image synthesis, both qualitatively and quantitatively.
\begin{table*}[h]
\centering
\scalebox{0.8}{
    \begin{tabular}{lc|cc|cc|cc}
    \multicolumn{2}{c}{} & \multicolumn{2}{c}{{\bf No Supervision}} & \multicolumn{2}{c}{{\bf Full Supervision}} & \multicolumn{2}{c}{{\bf Weak Supervision}} \\
    \hline
    Dataset & Real & CycleGAN & BiGAN & Disc-CGAN & Cont-CGAN & DFI & PC-GAN \\
    \hline
    CACD     & ~~94.37(\small{train})~~49.00(\small{val}) & 20.52 & 19.66 & 46.02 & 41.62 & 20.92 & 48.44 \\
    UTK      & ~~98.19(\small{train})~~76.80(\small{val}) & 19.46 & 20.50 & 71.44 & 59.16 & 22.90 & 63.88 \\
    SCUT-FBP &  100.00(\small{train})~~58.00(\small{val})  & 19.75 & 20.38 & 29.63 & 46.25 & 22.69 & 40.00 \\
    \hline
    Average Rank & -- & 5.67 & 5.33 & 2.00 & 2.33 & 4.00 & \textbf{1.67} \\
    \hline
    \end{tabular}}
    \caption{Evaluation of classification accuracies on synthesized images, higher is better.}
    \label{tab:baseline_acc}
\end{table*}
\begin{table*}[h]
    \centering
    \scalebox{0.8}{
    \begin{tabular}{llllccc|ccc}
    \multicolumn{4}{c}{{\bf Loss}} & \multicolumn{3}{c}{{\bf CACD}} & \multicolumn{3}{c}{{\bf UTKFace}} \\
    \hline
    \small{\texttt{CGAN}}  &\small{\texttt{rec}}   &\small{\texttt{cyc}}   &\small{\texttt{idt}}    & Acc (\%) & IS & FID & Acc (\%) & IS & FID \\ \hline
    \cmark&\cmark&\cmark&\cmark & 48.08 & 2.87$\pm$0.04 & 27.90$\pm$0.44 & 62.74 & 3.50$\pm$0.04 & 21.63$\pm$0.52 \\
    \cmark&\xmark&\cmark&\cmark & 39.50 & 2.93$\pm$0.04 & 25.68$\pm$0.46 & 56.90 & 3.38$\pm$0.05 & 24.98$\pm$0.88 \\
    \cmark&\cmark&\xmark&\cmark & 50.86 & 3.10$\pm$0.04 & 25.93$\pm$0.55 & 60.56 & 3.39$\pm$0.05 & 23.70$\pm$0.65 \\
    \cmark&\cmark&\cmark&\xmark & 48.60 & 3.05$\pm$0.03 & 26.81$\pm$0.59 & 63.92 & 3.60$\pm$0.05 & 27.65$\pm$0.75 \\
    \cmark&\cmark&\xmark&\xmark & 48.98 & 3.01$\pm$0.03 & 26.90$\pm$0.67 & 66.34 & 3.65$\pm$0.04 & 25.39$\pm$0.86 \\
    \cmark&\xmark&\cmark&\xmark & 24.28 & 3.06$\pm$0.04 & 24.01$\pm$0.66 & 50.42 & 3.02$\pm$0.04 & 48.80$\pm$1.70 \\
    \cmark&\xmark&\xmark&\cmark & 43.86 & 2.94$\pm$0.05 & 24.27$\pm$0.58 & 62.42 & 3.54$\pm$0.04 & 32.87$\pm$1.47 \\
    \cmark&\xmark&\xmark&\xmark & 20.08 & 1.59$\pm$0.02 & 293.03$\pm$1.40 & 34.88 & 2.16$\pm$0.04 & 187.98$\pm$2.17 \\
    \hline
    \end{tabular}}
    \caption{Ablation studies of different loss terms in CGAN training. {$\texttt{CGAN}$} represents {$\mathcal{L}_{CGAN}$}, {$\texttt{rec}$} represents {$\mathcal{L}_{rec}$} and so on.}
\label{tab:ablation}
\end{table*}
\begin{table*}[h]
    \centering
    \scalebox{0.8}{
    \begin{tabular}{lccc|ccc}
    \multicolumn{1}{c}{} & \multicolumn{3}{c}{{\bf CACD}} & \multicolumn{3}{c}{{\bf UTKFace}} \\
    \hline
    Model        & Acc (\%) & IS & FID & Acc (\%) & IS & FID \\ \hline
    CNN-CGAN          & 35.04 & 2.14$\pm$0.02 & 31.08$\pm$0.54 & 40.12 & 2.69$\pm$0.03 & 26.58$\pm$0.51 \\
    BNN-CGAN          & 37.64 & 2.38$\pm$0.04 & 27.36$\pm$0.36 & 38.54 & 2.72$\pm$0.03 & 26.56$\pm$0.40 \\
    BNN-RCGAN & 41.02 & 2.45$\pm$0.03 & 30.22$\pm$0.51 & 43.64 & 2.84$\pm$0.04 & 25.25$\pm$0.39 \\
    \hline
    \end{tabular}}
    \caption{Ablation study of Bayesian uncertainty estimation. \texttt{CNN-CGAN} is the normal non-Bayesian Elo rating network without uncertainty estimations; \texttt{BNN-CGAN} uses the average ratings for a single image; \texttt{BNN-RCGAN} is the full Bayesian model with a noise-robust CGAN.}
    \vspace{-2mm}
\label{tab:bnn}
\end{table*}

\noindent\textbf{Dataset.} We evaluate PC-GAN on a variety of datasets for image attribute editing tasks:
\begin{itemize}
    \item Annotated MNIST~\cite{kim2017mnist} provides annotations of stroke thickness for MNIST~\cite{lecun1998gradient} dataset.
    \item CACD~\cite{chen2014cross} is a large dataset collected for cross-age face recognition, which includes 2,000 subjects and 163,446 images. It contains multiple images for each person which cover different ages.
    \item UTKFace~\cite{zhifei2017cvpr} is also a large-scale face dataset with a long age span, ranging from 0 to 116 years. This dataset contains 23,709 facial images with annotations of age, gender, and ethnicity.
    \item SCUT-FBP~\cite{xie2015scut} is specifically designed for facial beauty perception. It contains 500 Asian female portraits with attractiveness ratings (1 to 5) labeled by 75 human raters.
    \item CelebA~\cite{liu2015faceattributes} is a standard large-scale dataset for facial attribute editing. It consists of over 200k images, annotated with 40 binary attributes.
\end{itemize}
For the MNIST experiment, stroke thickness is the desired attribute. As illustrated in Figure~\ref{fig:mnist}-a, the thickness information is still entangled. But in Figure~\ref{fig:mnist}-b, the thickness is correctly disentangled from the rest attributes.

We use CACD and UTK for age progression, SCUT-FBP and CelebA for attractiveness experiment. Since no true relatively labeled dataset is publically available, pairs are {\em simulated} from ``ground-truth'' attribute intensity given in the dataset. The tie margins within which two candidates are considered equal are 10, 10, and 0.4 for CACD, UTK, and SCUT-FBP, respectively. This also simplifies the quantitative evaluation process since one can directly measure the prediction error for absolute attribute intensities. Notice that CelebA only provides binary annotations, from which pairwise comparisons are simulated. Interestingly, the Elo rating network is still able to recover approximate ratings from those binary labels.

Furthermore, since CACD, UTKFace, SCUT-FBP, and CelebA are all human face dataset, we add an identity preserving loss term ~\cite{wang2018face} to enforce identity preservation: $\mathcal{L}_{idt} = \mathbb{E}_{x\sim p(x), y\sim p(y)}{\Vert h(\mathcal{G}(x,y))-h(x)\Vert^2_2}$. Here, $h(\cdot)$ denotes a pre-trained convnet.

\noindent \textbf{Implementation.}
PC-GAN is implemented using PyTorch~\cite{paszke2017automatic}. Network architectures and training details are given in Supplementary. For a fair evaluation, the basic modules are kept identical across all baselines.

\subsection{Learning by Pairwise Comparison}
\begin{figure}[h]
  \begin{center}
    \subfloat[CACD]{\includegraphics[width=0.3\linewidth]{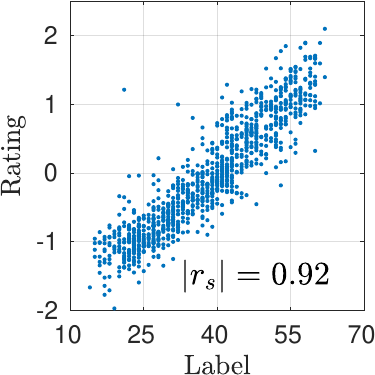}}
    \hspace{5pt}
    \subfloat[UTKFace]{\includegraphics[width=0.3\linewidth]{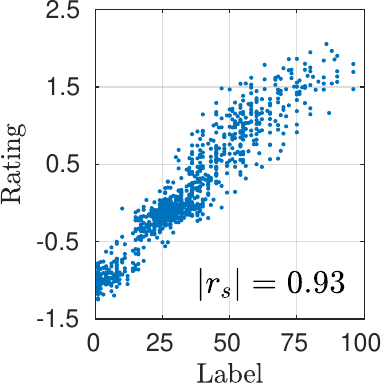}}
    \hspace{5pt}
    \subfloat[SCUT-FBP]{\includegraphics[width=0.3\linewidth]{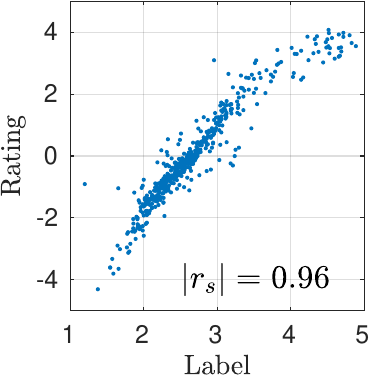}}
    \caption{Visualization of learned ratings for different datasets. $r_s$ denotes the Spearman's rank correlation coefficient.}
    \vspace{-2mm}
    \label{fig:emb_vis}
  \end{center}
\end{figure}

\begin{figure}[h]
  \begin{center}
    \subfloat[CACD]{\includegraphics[width=0.32\linewidth]{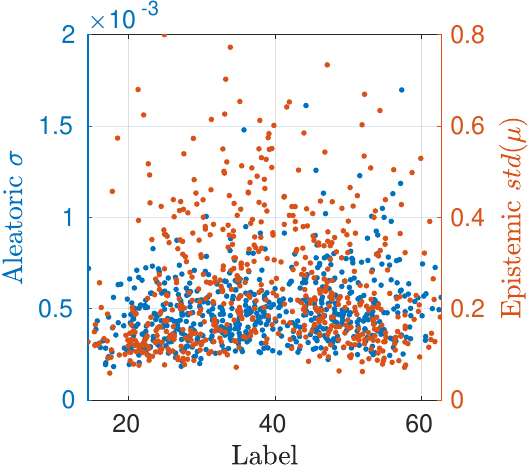}}
    \hspace{1pt}
    \subfloat[UTKFace]{\includegraphics[width=0.32\linewidth]{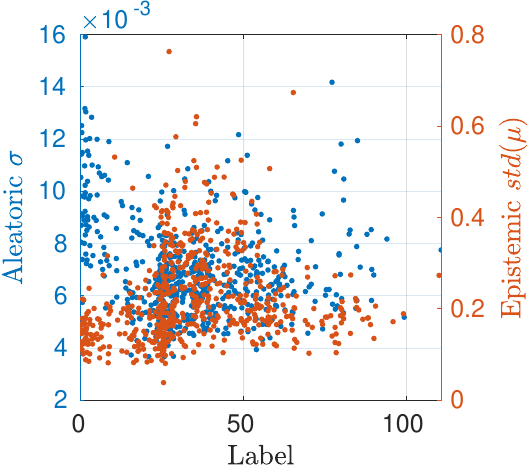}}
    \hspace{1pt}
    \subfloat[SCUT-FBP]{\includegraphics[width=0.32\linewidth]{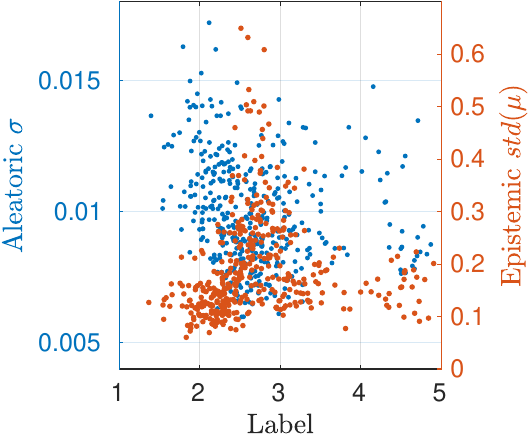}}
    \caption{Visualization of the predictive uncertainty of learned ratings for different datasets (best viewed in color). Aleatoric (data-dependent) and epistemic (model-dependent) uncertainties are plotted separately.}
    \vspace{-2mm}
    \label{fig:emb_std}
  \end{center}
\end{figure}

\noindent \textbf{Rating visualization.}
Figure~\ref{fig:emb_vis} presents the predicted ratings learned from CACD, UTKFace, and SCUT-FBP from left to right. The ratings learned from pairwise comparisons highly correlate with the ground-truth labels, which indicates that the rating resembles the attribute intensity well. The uncertainties v.s. ground-truth labels is visualized in Figure~\ref{fig:emb_std}. The plots show a general trend that the model is more certain about instances with extreme attribute values than those in the middle range, which matches our intuition. Additional attention-based visualizations are given in Supplementary.

\noindent \textbf{Noise resistance.} 
As mentioned previously, not only does pairwise comparison require less annotating effort, it tends to yield more accurate annotations. Consider a simple setting: if all annotators (annotating the absolute attribute value) exhibit the same random noise with a tie margin $M$, then the corresponding pairwise annotation with the same tie margin would absorb the noise. We provide an empirical study of the noise resistance of pairwise comparisons in Supplementary.

\subsection{Conditional Image Synthesis}
\noindent \textbf{Baselines.} We consider two unsupervised baselines CycleGAN and BiGAN, two fully-supervised baselines Disc-CGAN and Cont-CGAN, and DFI in a similar weakly-supervised setting.
\begin{itemize}
    \item \textbf{CycleGAN}~\cite{CycleGAN2017} learns an encoder (or a ``generator’’ from images to attributes) and a generator between images and attributes simultaneously.
    \item \textbf{ALI/BiGAN}\cite{donahue2016adversarial,dumoulin2016adversarially} learns the encoder (an inverse mapping) with a single discriminator.
    \item \textbf{Disc-CGAN/IPCGAN}~\cite{wang2018face} takes discretized attribute intensities (one-hot embedding) as supervision.
    \item \textbf{Cont-CGAN} uses the same CGAN framework as PC-GAN but ratings are replaced by true labels. It is an upper bound of PC-GAN.
    \item \textbf{DFI}~\cite{upchurch2017deep} can control the intensity of attribute intensity continuously, however, cannot change the intensity quantitatively. To transform $x$ into $\tilde{x}'$, we assume $\phi({\tilde{x}'})=\phi({x}) + \alpha {w}$ and compute $y'={w}\cdot\phi({x'})$ ($w$ is the attribute vector), then $\alpha$ is given by $\alpha = ({y'-{w}\cdot\phi({x})})/{\left\lVert {w}\right\rVert_2^2}$.
\end{itemize}

\noindent \textbf{Qualitative results.}
In Figure~\ref{fig:baselines}, we compare our results with all baselines. For each row, we take a source and a target image as inputs and our goal is to edit the attribute value of the source image to be equal to that of the target image. PC-GAN is competitive with fully-supervised baselines while all unsupervised methods fail to change attribute intensities.

More results are shown in Figure~\ref{fig:results_cacd}, \ref{fig:results_scut}, \ref{fig:results_utkface}, where the target rating value is the average of (cluster mean) a batch of (10 to 50) labeled images. From Figure~\ref{fig:results_cacd}, we see aging characteristics like receding hairlines and wrinkles are well learned. Figure~\ref{fig:results_utkface} shows convincing indications of rejuvenation and age progression. Figure~\ref{fig:results_scut} shows results for SCUT-FBP, which is inherently challenging because of the size of the dataset. Compared to datasets such as CACD, SCUT-FBP is significantly smaller, with only 500 images in total (from which we take 400 for training). Training on large datasets, as the CelebA experiment in Figure~\ref{fig:results_celeba} shows, our model produces convincing results. We also find that the model is capable of learning important patterns that correspond to attractiveness, such as in the hairstyle and the shape of the cheek  shown in Figure~\ref{fig:results_scut}. (The result does not represent the authors' opinion of attractiveness, but only reflects the statistics of the annotations.)

\noindent \textbf{Quantitative results.}
For quantitative evaluations, we report in Table~\ref{tab:baseline_acc} classification accuracy (Acc) evaluated on synthesized images. In our experiments, we train classifiers to predict attribute intensities of images into discrete groups (CACD $11\mhyphen20$, $21\mhyphen30$, up to $>50$; UTK $1\mhyphen20$, $21\mhyphen40$, up to $>80$, SCUT-FBP $1\mhyphen1.75$, $1.75\mhyphen2.5$, up to $>4$). PC-GAN demonstrates comparable performance with fully-supervised baselines and are significantly better than unsupervised methods. Additional metrics are reported in the Supplementary.

\noindent \textbf{AMT user studies.}
We also conduct user study experiments. Workers from Amazon Mechanical Turk (AMT) are asked to rate the quality of each face (good or bad) and vote to which age group a given image belongs. Then we calculate the percentage of images rated as good and the classification accuracy. Table~\ref{tab:user} shows that PC-GAN is on a par with the fully-supervised counterparts. We conduct hypothesis testing of PC-GAN and Disc-CGAN for image quality rating, $p\mhyphen\text{value}=0.31$, which indicates they are not statistically different with $95\%$ confidence level.
\begin{table}[h]
    \centering
    \scalebox{0.8}{
    \begin{tabular}{lcc|cc}
    \multicolumn{1}{c}{} & \multicolumn{2}{c}{{\bf CACD}} & \multicolumn{2}{c}{{\bf UTKFace}} \\
    \hline
    Method   & Quality (\%) & Acc (\%) & Quality (\%) & Acc (\%) \\
    \hline
    Real      &   97 &  36 &   88 &  52 \\
    PC-GAN    &   57 &  33 &   56 &  50 \\
    Cont-CGAN &   60 &  31 &   55 &  37 \\
    Disc-CGAN &   64 &  30 &   54 &  45 \\
    \hline
    \end{tabular}}
    \caption{AMT user studies. 100 images are sampled uniformly for each method with 20 images in each group.}
    \label{tab:user}
\end{table}

\subsection{Ablation Studies}
\noindent \textbf{Supervision.} First, the comparisons in Table~\ref{tab:baseline_acc} serve as an ablation study of full, no, and weak supervision, where PC-GAN is on a par with fully-supervised and significantly better than unsupervised baselines.

\noindent \textbf{GAN loss terms.} Second, an ablation study of CGAN loss terms is provided in Table~\ref{tab:ablation}. Notice that setting some losses to zero is a special case of our full objective under different $\lambda$s. Although we did not extensively tune $\lambda$’s values since it is not the main focus of this paper, we conclude that $\mathcal{L}_{rec}$ is the most important term in terms of image qualities.

\noindent \textbf{Uncertainty.} The ablation study of the effectiveness of adding Bayesian uncertainties to achieve robust conditional adversarial training is given in Table~\ref{tab:bnn}. The three variants considered in the table differ in how much the Bayesian neural net is involved in the whole training pipeline: \texttt{CNN-CGAN} is a non-Bayesian Elo rating network plus a normal CGAN, \texttt{BNN-CGAN} learns a Bayesian encoder and yields the average ratings for a given image, and \texttt{BNN-RCGAN} trains a full Bayesian encoder with a noise-robust CGAN. Results confirm that the performance can be boosted by integrating an uncertainty-aware Elo rating network and an extended robust conditional GAN.

\section{Conclusion}
In this paper, we propose a noise-robust conditional GAN framework under weak supervision for image attribute editing. Our method can learn an attribute rating function and estimate the predictive uncertainties from pairwise comparisons, which requires less annotation effort. We show in extensive experiments that the proposed PC-GAN performs competitively with the supervised baselines and significantly outperforms the unsupervised baselines.


\subsection{Acknowledgments}
We would like to thank Fei Deng for valuable discussions on Elo rating networks. This research is supported in part by NSF 1763523, 1747778, 1733843, and 1703883.

\bibliographystyle{aaai}
\bibliography{ref}

\begin{thebibliography}{}

\bibitem[\protect\citeauthoryear{Bora, Price, and
  Dimakis}{2018}]{bora2018ambientgan}
Bora, A.; Price, E.; and Dimakis, A.~G.
\newblock 2018.
\newblock Ambientgan: Generative models from lossy measurements.
\newblock {\em ICLR} 2:5.

\bibitem[\protect\citeauthoryear{Burges \bgroup et al\mbox.\egroup
  }{2005}]{burges2005learning}
Burges, C.; Shaked, T.; Renshaw, E.; Lazier, A.; Deeds, M.; Hamilton, N.; and
  Hullender, G.
\newblock 2005.
\newblock Learning to rank using gradient descent.
\newblock In {\em Proceedings of the 22nd international conference on Machine
  learning},  89--96.
\newblock ACM.

\bibitem[\protect\citeauthoryear{Chen \bgroup et al\mbox.\egroup
  }{2016}]{chen2016infogan}
Chen, X.; Duan, Y.; Houthooft, R.; Schulman, J.; Sutskever, I.; and Abbeel, P.
\newblock 2016.
\newblock Infogan: Interpretable representation learning by information
  maximizing generative adversarial nets.
\newblock In {\em Advances in neural information processing systems},
  2172--2180.

\bibitem[\protect\citeauthoryear{Chen, Chen, and Hsu}{2014}]{chen2014cross}
Chen, B.-C.; Chen, C.-S.; and Hsu, W.~H.
\newblock 2014.
\newblock Cross-age reference coding for age-invariant face recognition and
  retrieval.
\newblock In {\em European conference on computer vision},  768--783.
\newblock Springer.

\bibitem[\protect\citeauthoryear{Donahue, Kr{\"a}henb{\"u}hl, and
  Darrell}{2016}]{donahue2016adversarial}
Donahue, J.; Kr{\"a}henb{\"u}hl, P.; and Darrell, T.
\newblock 2016.
\newblock Adversarial feature learning.
\newblock {\em arXiv preprint arXiv:1605.09782}.

\bibitem[\protect\citeauthoryear{Dumoulin \bgroup et al\mbox.\egroup
  }{2016}]{dumoulin2016adversarially}
Dumoulin, V.; Belghazi, I.; Poole, B.; Mastropietro, O.; Lamb, A.; Arjovsky,
  M.; and Courville, A.
\newblock 2016.
\newblock Adversarially learned inference.
\newblock {\em arXiv preprint arXiv:1606.00704}.

\bibitem[\protect\citeauthoryear{Elo}{1978}]{elo1978rating}
Elo, A.~E.
\newblock 1978.
\newblock {\em The rating of chessplayers, past and present}.
\newblock Arco Pub.

\bibitem[\protect\citeauthoryear{F{\"u}rnkranz and
  H{\"u}llermeier}{2010}]{furnkranz2010preference}
F{\"u}rnkranz, J., and H{\"u}llermeier, E.
\newblock 2010.
\newblock Preference learning and ranking by pairwise comparison.
\newblock In {\em Preference learning}. Springer.
\newblock  65--82.

\bibitem[\protect\citeauthoryear{Gal and Ghahramani}{2016}]{gal2016dropout}
Gal, Y., and Ghahramani, Z.
\newblock 2016.
\newblock Dropout as a bayesian approximation: Representing model uncertainty
  in deep learning.
\newblock In {\em international conference on machine learning},  1050--1059.

\bibitem[\protect\citeauthoryear{Goodfellow \bgroup et al\mbox.\egroup
  }{2014}]{goodfellow2014generative}
Goodfellow, I.; Pouget-Abadie, J.; Mirza, M.; Xu, B.; Warde-Farley, D.; Ozair,
  S.; Courville, A.; and Bengio, Y.
\newblock 2014.
\newblock Generative adversarial nets.
\newblock In {\em Advances in neural information processing systems},
  2672--2680.

\bibitem[\protect\citeauthoryear{Han, Murphy, and
  Ramanan}{2018}]{han2018learning}
Han, L.; Murphy, R.~F.; and Ramanan, D.
\newblock 2018.
\newblock Learning generative models of tissue organization with supervised
  gans.
\newblock In {\em 2018 IEEE Winter Conference on Applications of Computer
  Vision (WACV)},  682--690.
\newblock IEEE.

\bibitem[\protect\citeauthoryear{He \bgroup et al\mbox.\egroup
  }{2016}]{he2016deep}
He, K.; Zhang, X.; Ren, S.; and Sun, J.
\newblock 2016.
\newblock Deep residual learning for image recognition.
\newblock In {\em Proceedings of the IEEE conference on computer vision and
  pattern recognition},  770--778.

\bibitem[\protect\citeauthoryear{Herbrich, Minka, and
  Graepel}{2007}]{herbrich2007trueskill}
Herbrich, R.; Minka, T.; and Graepel, T.
\newblock 2007.
\newblock Trueskill™: a bayesian skill rating system.
\newblock In {\em Advances in neural information processing systems},
  569--576.

\bibitem[\protect\citeauthoryear{Heusel \bgroup et al\mbox.\egroup
  }{2017}]{heusel2017gans}
Heusel, M.; Ramsauer, H.; Unterthiner, T.; Nessler, B.; and Hochreiter, S.
\newblock 2017.
\newblock Gans trained by a two time-scale update rule converge to a local nash
  equilibrium.
\newblock In {\em Advances in Neural Information Processing Systems},
  6626--6637.

\bibitem[\protect\citeauthoryear{Kendall and
  Gal}{2017}]{kendall2017uncertainties}
Kendall, A., and Gal, Y.
\newblock 2017.
\newblock What uncertainties do we need in bayesian deep learning for computer
  vision?
\newblock In {\em Advances in neural information processing systems},
  5574--5584.

\bibitem[\protect\citeauthoryear{Kim}{2017}]{kim2017mnist}
Kim, B.
\newblock 2017.
\newblock Annotated mnist: Thickness and skew labeler for mnist handwritten
  digit dataset.
\newblock https://github.com/1202kbs/Annotated\_MNIST.

\bibitem[\protect\citeauthoryear{Kingma and Welling}{2013}]{kingma2013auto}
Kingma, D.~P., and Welling, M.
\newblock 2013.
\newblock Auto-encoding variational bayes.
\newblock {\em arXiv preprint arXiv:1312.6114}.

\bibitem[\protect\citeauthoryear{LeCun \bgroup et al\mbox.\egroup
  }{1998}]{lecun1998gradient}
LeCun, Y.; Bottou, L.; Bengio, Y.; Haffner, P.; et~al.
\newblock 1998.
\newblock Gradient-based learning applied to document recognition.
\newblock {\em Proceedings of the IEEE} 86(11):2278--2324.

\bibitem[\protect\citeauthoryear{Liu \bgroup et al\mbox.\egroup
  }{2015}]{liu2015faceattributes}
Liu, Z.; Luo, P.; Wang, X.; and Tang, X.
\newblock 2015.
\newblock Deep learning face attributes in the wild.
\newblock In {\em Proceedings of International Conference on Computer Vision
  (ICCV)}.

\bibitem[\protect\citeauthoryear{Lucic \bgroup et al\mbox.\egroup
  }{2019}]{lucic2019high}
Lucic, M.; Tschannen, M.; Ritter, M.; Zhai, X.; Bachem, O.; and Gelly, S.
\newblock 2019.
\newblock High-fidelity image generation with fewer labels.
\newblock {\em arXiv preprint arXiv:1903.02271}.

\bibitem[\protect\citeauthoryear{Mirza and
  Osindero}{2014}]{mirza2014conditional}
Mirza, M., and Osindero, S.
\newblock 2014.
\newblock Conditional generative adversarial nets.
\newblock {\em arXiv preprint arXiv:1411.1784}.

\bibitem[\protect\citeauthoryear{Negahban, Oh, and
  Shah}{2016}]{negahban2016rank}
Negahban, S.; Oh, S.; and Shah, D.
\newblock 2016.
\newblock Rank centrality: Ranking from pairwise comparisons.
\newblock {\em Operations Research} 65(1):266--287.

\bibitem[\protect\citeauthoryear{Odena, Olah, and
  Shlens}{2016}]{odena2016conditional}
Odena, A.; Olah, C.; and Shlens, J.
\newblock 2016.
\newblock Conditional image synthesis with auxiliary classifier gans.
\newblock {\em arXiv preprint arXiv:1610.09585}.

\bibitem[\protect\citeauthoryear{Paszke \bgroup et al\mbox.\egroup
  }{2017}]{paszke2017automatic}
Paszke, A.; Gross, S.; Chintala, S.; Chanan, G.; Yang, E.; DeVito, Z.; Lin, Z.;
  Desmaison, A.; Antiga, L.; and Lerer, A.
\newblock 2017.
\newblock Automatic differentiation in pytorch.

\bibitem[\protect\citeauthoryear{Radinsky and
  Ailon}{2011}]{radinsky2011ranking}
Radinsky, K., and Ailon, N.
\newblock 2011.
\newblock Ranking from pairs and triplets: information quality, evaluation
  methods and query complexity.
\newblock In {\em Proceedings of the fourth ACM international conference on Web
  search and data mining},  105--114.
\newblock ACM.

\bibitem[\protect\citeauthoryear{Salimans \bgroup et al\mbox.\egroup
  }{2016}]{salimans2016improved}
Salimans, T.; Goodfellow, I.; Zaremba, W.; Cheung, V.; Radford, A.; and Chen,
  X.
\newblock 2016.
\newblock Improved techniques for training gans.
\newblock In {\em Advances in Neural Information Processing Systems},
  2234--2242.

\bibitem[\protect\citeauthoryear{Selvaraju \bgroup et al\mbox.\egroup
  }{2017}]{selvaraju2017grad}
Selvaraju, R.~R.; Cogswell, M.; Das, A.; Vedantam, R.; Parikh, D.; and Batra,
  D.
\newblock 2017.
\newblock Grad-cam: Visual explanations from deep networks via gradient-based
  localization.
\newblock In {\em Proceedings of the IEEE International Conference on Computer
  Vision},  618--626.

\bibitem[\protect\citeauthoryear{Shrivastava, Gupta, and
  Girshick}{2016}]{shrivastava2016training}
Shrivastava, A.; Gupta, A.; and Girshick, R.
\newblock 2016.
\newblock Training region-based object detectors with online hard example
  mining.
\newblock In {\em Proceedings of the IEEE Conference on Computer Vision and
  Pattern Recognition},  761--769.

\bibitem[\protect\citeauthoryear{Thekumparampil \bgroup et al\mbox.\egroup
  }{2018}]{thekumparampil2018robustness}
Thekumparampil, K.~K.; Khetan, A.; Lin, Z.; and Oh, S.
\newblock 2018.
\newblock Robustness of conditional gans to noisy labels.
\newblock In {\em Advances in Neural Information Processing Systems},
  10271--10282.

\bibitem[\protect\citeauthoryear{Upchurch \bgroup et al\mbox.\egroup
  }{2017}]{upchurch2017deep}
Upchurch, P.; Gardner, J.~R.; Pleiss, G.; Pless, R.; Snavely, N.; Bala, K.; and
  Weinberger, K.~Q.
\newblock 2017.
\newblock Deep feature interpolation for image content changes.
\newblock In {\em CVPR},  6090--6099.

\bibitem[\protect\citeauthoryear{Wang \bgroup et al\mbox.\egroup
  }{2018a}]{wang2018weakly}
Wang, Y.; Wang, S.; Qi, G.; Tang, J.; and Li, B.
\newblock 2018a.
\newblock Weakly supervised facial attribute manipulation via deep adversarial
  network.
\newblock In {\em 2018 IEEE Winter Conference on Applications of Computer
  Vision (WACV)},  112--121.
\newblock IEEE.

\bibitem[\protect\citeauthoryear{Wang \bgroup et al\mbox.\egroup
  }{2018b}]{wang2018face}
Wang, Z.; Tang, X.; Luo, W.; and Gao, S.
\newblock 2018b.
\newblock Face aging with identity-preserved conditional generative adversarial
  networks.
\newblock In {\em Proceedings of the IEEE Conference on Computer Vision and
  Pattern Recognition},  7939--7947.

\bibitem[\protect\citeauthoryear{Wauthier, Jordan, and
  Jojic}{2013}]{wauthier2013efficient}
Wauthier, F.; Jordan, M.; and Jojic, N.
\newblock 2013.
\newblock Efficient ranking from pairwise comparisons.
\newblock In {\em International Conference on Machine Learning},  109--117.

\bibitem[\protect\citeauthoryear{Xiao and Jae~Lee}{2015}]{xiao2015discovering}
Xiao, F., and Jae~Lee, Y.
\newblock 2015.
\newblock Discovering the spatial extent of relative attributes.
\newblock In {\em Proceedings of the IEEE International Conference on Computer
  Vision},  1458--1466.

\bibitem[\protect\citeauthoryear{Xie \bgroup et al\mbox.\egroup
  }{2015}]{xie2015scut}
Xie, D.; Liang, L.; Jin, L.; Xu, J.; and Li, M.
\newblock 2015.
\newblock Scut-fbp: A benchmark dataset for facial beauty perception.
\newblock {\em arXiv preprint arXiv:1511.02459}.

\bibitem[\protect\citeauthoryear{Xu \bgroup et al\mbox.\egroup
  }{2019}]{xu2019deep}
Xu, Q.; Yang, Z.; Jiang, Y.; Cao, X.; Huang, Q.; and Yao, Y.
\newblock 2019.
\newblock Deep robust subjective visual property prediction in crowdsourcing.
\newblock In {\em Proceedings of the IEEE Conference on Computer Vision and
  Pattern Recognition},  8993--9001.

\bibitem[\protect\citeauthoryear{Yan}{2016}]{yanpassive}
Yan, S.
\newblock 2016.
\newblock Passive and active ranking from pairwise comparisons.
\newblock Technical report, University of California, San Diego.

\bibitem[\protect\citeauthoryear{Zhang and Qi}{2017}]{zhifei2017cvpr}
Zhang, Zhifei, S.~Y., and Qi, H.
\newblock 2017.
\newblock Age progression/regression by conditional adversarial autoencoder.
\newblock In {\em IEEE Conference on Computer Vision and Pattern Recognition
  (CVPR)}.
\newblock IEEE.

\bibitem[\protect\citeauthoryear{Zhou}{2017}]{zhou2017brief}
Zhou, Z.-H.
\newblock 2017.
\newblock A brief introduction to weakly supervised learning.
\newblock {\em National Science Review} 5(1):44--53.

\bibitem[\protect\citeauthoryear{Zhu \bgroup et al\mbox.\egroup
  }{2017}]{CycleGAN2017}
Zhu, J.-Y.; Park, T.; Isola, P.; and Efros, A.~A.
\newblock 2017.
\newblock Unpaired image-to-image translation using cycle-consistent
  adversarial networks.
\newblock In {\em Proceedings of the IEEE international conference on computer
  vision},  2223--2232.

\end{thebibliography}

\appendix

\section{Supplementary}

In Supplementary, we first show the analysis of CGAN loss terms and give a proof of {Proposition 0.1}. Then we provide an empirical study of how the number of pairs varies with the size of the dataset. The preliminary results on noise resistance is also presented. Next, we show qualitative attention visualization of the Elo rating network and report additional quantitative IS and FID scores for baselines and list details of network architectures. Finally, we show additional results on conditional image synthesis.

\section{Analysis of Loss Terms}
As a standard recall in~\cite{goodfellow2014generative}, the adversarial training results in minimizing the Jensen-Shannon divergence between the true conditional and the generated conditional. We show that the following proposition holds:
\begin{theorem}
The global minimum of $\mathcal{L}(\mathcal{G}, \mathcal{D})$ is achieved if and only if $q_{\mathcal{G}}(\tilde{x}'|x, y') = p_{\mathcal{E}}(\tilde{x}'|x, y')$, where $p$ is the true distribution and $q_{\mathcal{G}}$ is the distribution induced by $\mathcal{G}$.
\end{theorem}
\begin{proof}
$(x',y')$ is sampled from true distribution, $x$ is independently sampled and  $\tilde{x}'$ is sampled from Generator $G(x,y')$, rewrite Equation {\color{red}5} in integral form,

\begin{align}
\mathcal{L}_{CGAN} &= \int{p_{\mathcal{E}}(x', y') \log(\mathcal{D}(x', y')) dx' dy'} + \nonumber \\
& \int{p(x)p_{\mathcal{E}}(y')q_{\mathcal{G}}(\tilde{x}'|x, y')\log(1-\mathcal{D}(\tilde{x}', y')) dx dy' d\tilde{x}'} \nonumber \\
&= \int{p_{\mathcal{E}}(x, \tilde{x}', y') \log(\mathcal{D}(\tilde{x}', y'))} + \\
& {p_{\mathcal{E}}(x, y')q_{\mathcal{G}}(\tilde{x}'|x, y')\log(1-\mathcal{D}(\tilde{x}', y')) dx dy' d\tilde{x}'} \nonumber,
\end{align}
\noindent where we assume $x$ and $y'$ are sampled independently. We get the optimal discriminator $\mathcal{D}^*$ by applying Euler-Lagrange equation,
\begin{align}
\mathcal{D}^* &= \frac{p_{\mathcal{E}}(\tilde{x}' | x, y')}{p_{\mathcal{E}}(\tilde{x}' | x, y')+q_{\mathcal{G}}(\tilde{x}' | x, y')}.
\end{align}
\noindent Finally plugging $\mathcal{D}^*$ in $\mathcal{L}_{CGAN}$ yields,
\begin{align}
\mathcal{L}_{CGAN}&(\mathcal{G}, \mathcal{D}^*) = -2\log2 \quad + \\
&2\int p_{\mathcal{E}}(x, y')\text{JSD}{\left(p_{\mathcal{E}}(\tilde{x}' | x, y')||q_{\mathcal{G}}(\tilde{x}' | x, y')\right) dx dy'}, \nonumber
\end{align}
\noindent where $\text{JSD}$ is the Jensen-Shannon divergence. Since $\text{JSD}$ is always non-negative and reaches its minimum if and only if $q_{\mathcal{G}}(\tilde{x}' | x, y')=p_{\mathcal{E}}(\tilde{x}' | x, y')$ for $(x, y') \in \{(x,y'): p_{\mathcal{E}}(x, y')>0\}$, $\mathcal{G}$ recovers the true conditional distribution $p_{\mathcal{E}}(\tilde{x}' | x, y')$ when $\mathcal{D}$ and $\mathcal{G}$ are trained optimally.

In addition, the reconstruction loss $\mathcal{L}_{rec}^y$, cycle loss $\mathcal{L}_{cyc}$, and identity preserving loss $\mathcal{L}_{idt}$ are all non-negative. Minimizing these losses will keep the equilibrium of $\mathcal{L}_{CGAN}$. If the encoder $p_{\mathcal{E}}(y|x)$ and the feature extractor $h(\cdot)$ are trained properly, $\mathcal{L}(\mathcal{G}, \mathcal{D}^*)$ achieves its minimum when $\mathcal{G}$ is optimally trained.
\end{proof}

\section{Proof of Proposition 0.1}
\begin{proof}

For $\forall u,v \in V$, we define $\pi(u,v) = 1$ if $u < v$ and 0 otherwise, $w(u,v)$ measures the extent to which $u$ should be prefered over $v$, 

For any pair $u,v$, let 
\begin{align}
    L_{u,v} = \pi(u,v)w(u,v) + \pi(v,u)w(v,u)
\end{align}
where $\pi(u,v)$ is the ground-truth and $w(v,u)$ is prediction from Elo ranking network.

Define 
\begin{align}
L = \Sigma_{u<v, u,v\in V}L_{u,v}
\end{align}
 as our loss function and from results in \cite{radinsky2011ranking}, we have the lemma:
\begin{lemma}
For $\delta >0$, any $0<\lambda<1$, if we sample $dn/\lambda^2$ pairs uniformly with repetition from ${V}\choose{2}$, with probability $1 - \delta$, 
\begin{align}
    L(V,w,\hat{\pi}) \leq \lambda \Bigg[ \frac{c}{\sqrt{d}} + \sqrt{\frac{\log{\frac{1}{\delta}}}{dn}}\Bigg{]} {{n}\choose{2}}.
\end{align}
\end{lemma}

Define 
\begin{align}
\label{eqn:tdefintion}
    t = \lambda \Bigg[ \frac{c}{\sqrt{d}} + \sqrt{\frac{\log{\frac{1}{\delta}}}{dn}}\Bigg{]} {{n}\choose{2}},
\end{align}
and let $\delta = 1$, we get $t_1$ and $\mathbb{P}(L(\hat{\pi})>t_1)\leq\delta =1$
\begin{align}
    t_1 = \lambda \Bigg[ \frac{c}{\sqrt{d}} + \sqrt{\frac{\log{1}}{dn}}\Bigg{]} {{n}\choose{2}}.
\end{align}
\begin{align}
    \label{eqn:integral}
    \mathbb{E}(L(\hat{\pi})) = \int_0^{\infty}\mathbb{P}(L(\hat{\pi})>t)dt \leq t_1 + \int_{t_1}^{\infty}\mathbb{P}(L(\hat{\pi})>t)dt
\end{align}
From Equation~\ref{eqn:tdefintion}, 
\begin{align}
    \delta = exp(-\frac{1}2\sigma_n^2(t-\mu_n)^2)
\end{align}
where $\sigma_n^2 = \frac{\lambda^2(n(n-1))^2}{4dn}$, $\mu_n = \frac{\lambda n(n-1)c}{2\sqrt{d}}$.

Plugging back in Equation~\ref{eqn:integral}, 
\begin{align}
    \mathbb{E}(L(\hat{\pi}) \leq t_1 + \sqrt{2\pi\sigma_n^2} \nonumber
\end{align}
\begin{align}
   & = \lambda \Bigg[ \frac{c}{\sqrt{d}} + \sqrt{\frac{\log{1}}{dn}}\Bigg{]} {{n}\choose{2}} + 2\lambda\sqrt{2\pi}\frac{n(n-1)}{s\sqrt{dn}} \nonumber\\ 
   & = \lambda \Bigg[ \frac{c}{\sqrt{d}} + \frac{\sqrt{\log{1}}+\sqrt{8\pi}}{\sqrt{dn}}\Bigg{]} {{n}\choose{2}}.
\end{align}
Set $d = 16c^2$, for $\lambda/4 > \epsilon_0>0$, there is $n_0$ so that if $n>n_0$,

\begin{align}
\mathbb{E}(L(\hat{\pi})) \leq (\lambda/4 + \epsilon_0){{n}\choose{2}} \leq \lambda/2{{n}\choose{2}}.
\end{align}
\end{proof}

\section{Number of Pairs}
\begin{figure}
  \begin{center}
    \subfloat[$n=100$\vspace{-1.mm}]{\includegraphics[width=0.3\linewidth]{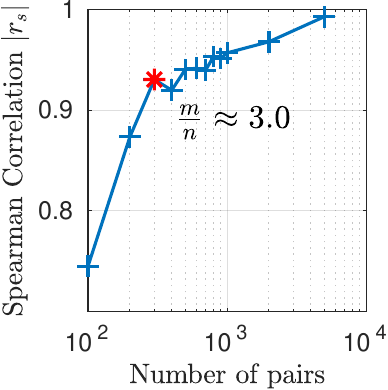}}
    \hspace{5pt}
    \subfloat[$n=500$\vspace{-1.mm}]{\includegraphics[width=0.3\linewidth]{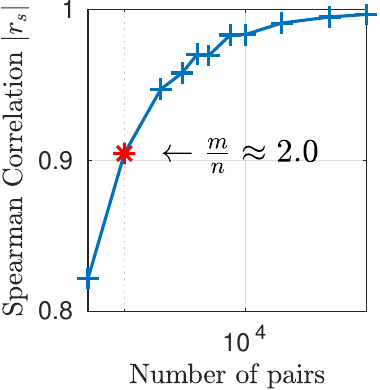}}
    \hspace{5pt}
    \subfloat[$n=1000$\vspace{-1.mm}]{\includegraphics[width=0.3\linewidth]{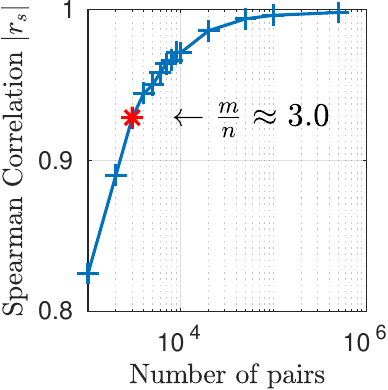}}
    \par
    \subfloat[$n=2000$\vspace{-1.mm}]{\includegraphics[width=0.3\linewidth]{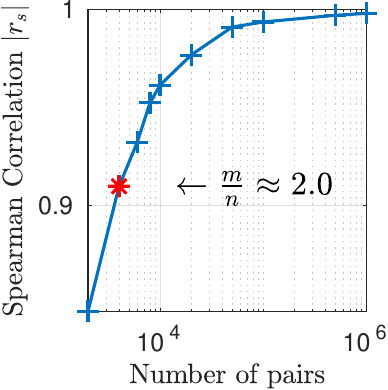}}
    \hspace{5pt}
    \subfloat[$n=5000$\vspace{-1.mm}]{\includegraphics[width=0.3\linewidth]{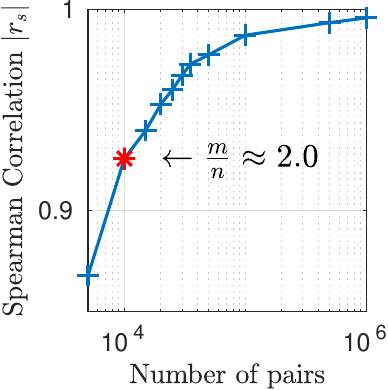}}
    \hspace{5pt}
    \subfloat[$n=10000$\vspace{-1.mm}]{\includegraphics[width=0.3\linewidth]{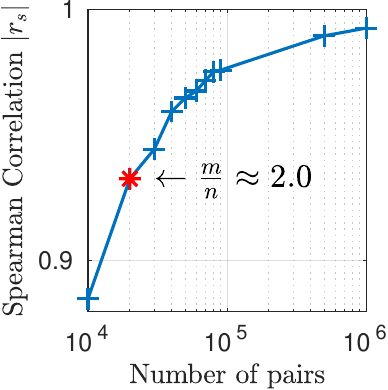}}
    \caption{Number of pairs $m$ v.s. Spearman correlation $r_s$. Different subsets of images (of number $n=100, \ldots, 10000$) are randomly selected from the UTKFace dataset. For each subset, different number of pairs (denoted by $m$) are randomly sampled. The smallest number of pairs with a Spearman's rank correlation coefficient that exceeds $0.9$ is marked by a red asterisk symbol {\color{red}\textbf{$\ast$}}. To achieve high correlations between ratings and labels (in terms of $|r_s| \geq 0.9$), approximately $2 n$ pairs are required.}
    \label{fig:npair}
  \end{center}
\end{figure}

To experimentally verify the number of pairs needed to learn a rating, we sampled from UTKFace~\cite{zhifei2017cvpr} subsets of sizes $100$, $500$, $1000$, $2000$, $5000$ and $10000$, and train Elo rating networks with different number of pairs for each subset. As illustrated in Figure~\ref{fig:npair}, to achieve a Spearman correlation above $0.9$, approximately $2n$ pairs are needed, where $n$ is the size of the subset. $n\log{n}$ comparisons are needed for exact recovery of ranking between $n$ objects. Through our ranking network, we need $\mathcal{O}(n)$ comparisons to learn rating that is close enough to the true attribute strength and also keeping the space between objects. Annotation of absolute attribute strength is very noisy and usually takes $\mathcal{O}(n)$ annotations because of majority voting (e.g. $3n$ if 3 workers per instance), our method doesn't require more effort in annotation and pairwise comparisons are easier to annotate comparing to absolute attribute strength, which will lead to a faster finishing time in crowd-sourcing phase.

\section{Noise Resistance}
\begin{figure}[h]
  \begin{center}
    \subfloat[CACD]{\includegraphics[width=0.3\linewidth]{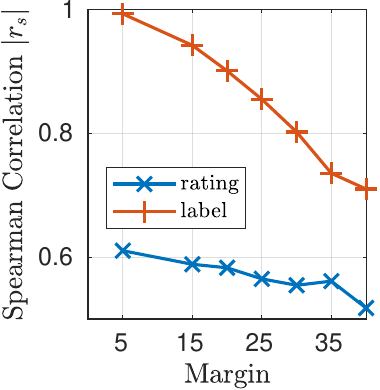}}
    \hspace{5pt}
    \subfloat[UTKFace]{\includegraphics[width=0.3\linewidth]{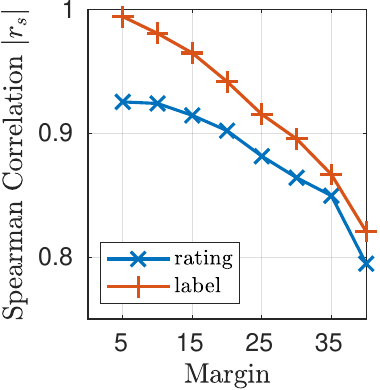}}
    \hspace{5pt}
    \subfloat[SCUT-FBP]{\includegraphics[width=0.3\linewidth]{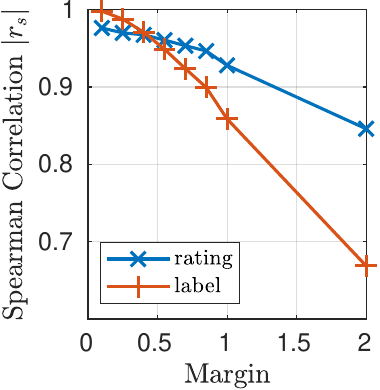}}
    \caption{Noise resistance. Spearman correlations between ground-truth labels and ratings or noisy labels under different tie margins (a tie margin is the range within which an agent is indifferent between two alternatives).}
    \label{fig:noise}
  \end{center}
\end{figure}

Considering there is noise when annotating the absolute labels. Taking age annotation as an example, we assume annotators will give $x$ an age $\Omega'(x)$ that deviates from the true age $\Omega(x)$ by a random noise: $\Omega'(x) = \Omega(x) + z, z \sim \text{Unif}\left(-\frac{M}{2},\frac{M}{2}\right)$, where $M$ is the tie margin in Figure~\ref{fig:noise}. As shown, the correlation curve of ratings drops slowly until the noise level is too high. Although only the curve on SCUT-FBP shows superior results over the ground-truth label, the general trend is that the rating curves decrease slower than the absolute label curves. This demonstrates the Elo rating network's potential of noise resistance.

We choose UTKFace dataset to investigate how conditional synthesis results might be affected by margins. In Table~\ref{tab:margin_utk}, Spearman correlations and Inception Scores evaluated on UTKFace under different margin values are reported.
\begin{table}[h]
    \centering
    \scalebox{0.8}{
    \begin{tabular}{lrrr}
    \hline
    Margin & {\bf Corr} &  {\bf Acc (\%)} & {\bf IS} \\
    \hline
    5  & 0.93 & 73.26 & 3.70$\pm$0.07 \\
    15 & 0.91 & 64.18 & 3.56$\pm$0.06 \\
    25 & 0.88 & 73.26 & 3.78$\pm$0.04 \\
    35 & 0.85 & 60.74 & 3.50$\pm$0.06 \\
    \hline
    \end{tabular}}
    \caption{Spearman correlations ({\bf Corr}), Inception Scores ({\bf IS}) evaluated on UTKFace under different margin values. Pairs are randomly sampled and CGANs are trained using different pairs.}
    \label{tab:margin_utk}
\end{table}

\section{Attention Visualization}
\begin{figure}[h]
  \vspace{-0.2in}
  \begin{center}
    \subfloat[Attention via Grad-CAM\vspace{-1mm}]{\includegraphics[scale=.375]{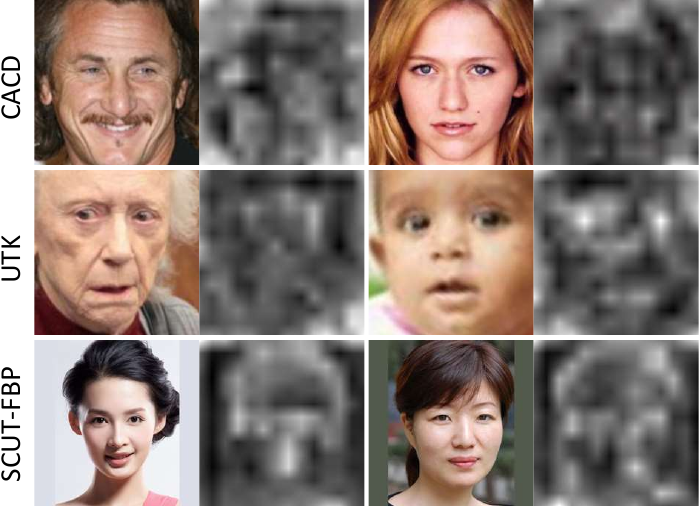}}
    \hspace{3pt}
    \subfloat[Inverting the encoder\vspace{-1mm}]{\includegraphics[scale=.375]{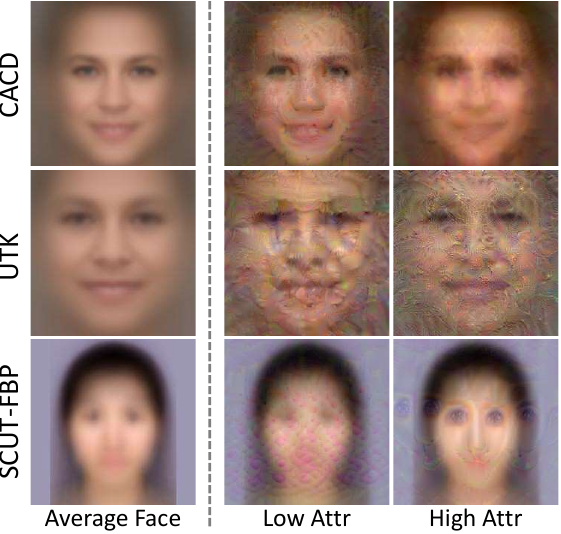}}
    \hspace{3pt}
    \caption{(a) Attention visualization for Elo rating network via Grad-CAM. (b) Inverting the Elo rating network by optimization over the input image (average faces) to match low/high attribute intensity.}
    \label{fig:emb_attn}
  \end{center}
  \vspace{-0.1in}
\end{figure}

The proposed Elo rating network is visualized using Grad-CAM~\cite{selvaraju2017grad}. In Figure~\ref{fig:emb_attn}-a, local regions that are critical for decision making are highlighted: for CACD and UTKFace, aging indicators such as forehead wrinkles, crow's feet eyes (babies usually have big eyes) are highlighted; for SCUT-FBP, the gradient map highlights facial regions like eyes, nose, pimples etc. Similar to DFI, if viewing the rating as deep features, one can optimize over the input image to obtain a new image with desired attribute intensity. We thus invert the encoders to see what a ``typical'' image with extreme attribute intensity would look like by optimizing the average face as shown in Figure~\ref{fig:emb_attn}-b.
\begin{table*}[t]
\centering
\subfloat[{\bf Inception Score} (higher is better)\vspace{-1.5mm}]{
    \scalebox{0.8}{
    \begin{tabular}{lr|rr|rr|rr}
    \multicolumn{2}{c}{} & \multicolumn{2}{c}{\small{weak supervision}} & \multicolumn{2}{c}{\small{full supervision}} & \multicolumn{2}{c}{\small{no supervision}} \\
    \hline
    Dataset & Real & PC-GAN & DFI & Cont-CGAN & Disc-CGAN & CycleGAN & BiGAN \\
    \hline
    CACD     & 3.89 $\pm$ 0.05 & 2.89 $\pm$ 0.06 & 3.35 $\pm$ 0.06 & 2.85 $\pm$ 0.03 & 2.95 $\pm$ 0.04 & 2.96 $\pm$ 0.03 & 3.27 $\pm$ 0.04 \\
    UTK      & 4.29 $\pm$ 0.05 & 3.55 $\pm$ 0.06 & 3.26 $\pm$ 0.06 & 3.52 $\pm$ 0.04 & 3.66 $\pm$ 0.04 & 3.09 $\pm$ 0.06 & 3.20 $\pm$ 0.06 \\
    SCUT-FBP & 4.20 $\pm$ 0.05 & 2.88 $\pm$ 0.11 & 2.93 $\pm$ 0.07 & 2.39 $\pm$ 0.14 & 1.37 $\pm$ 0.02 & 2.85 $\pm$ 0.15 & 3.05 $\pm$ 0.15 \\
    \hline
    \end{tabular}}
}\par
\subfloat[{\bf Fr\'{e}chet Inception Distance} (lower is better)\vspace{-1.5mm}]{
    \scalebox{0.8}{
    \begin{tabular}{lrr|rr|rr}
    \multicolumn{1}{c}{} & \multicolumn{2}{c}{\small{weak supervision}} & \multicolumn{2}{c}{\small{full supervision}} & \multicolumn{2}{c}{\small{no supervision}} \\
    \hline
    Dataset & PC-GAN & DFI & Cont-CGAN & Disc-CGAN & CycleGAN & BiGAN \\
    \hline
    CACD     & 28.20 $\pm$ 0.65 & 25.18 $\pm$ 0.73 & 28.53 $\pm$ 0.72 & 28.13 $\pm$ 0.71 & 26.76 $\pm$ 0.64 & 24.69 $\pm$ 0.62 \\
    UTK      & 24.86 $\pm$ 0.84 & 28.32 $\pm$ 0.75 & 28.42 $\pm$ 0.98 & 33.26 $\pm$ 1.49 & 23.16 $\pm$ 0.75 & 19.72 $\pm$ 0.79 \\
    SCUT-FBP & 97.21 $\pm$ 2.81 & 48.67 $\pm$ 1.42 & 114.89 $\pm$ 3.08 & 188.09 $\pm$ 3.91 & 87.07 $\pm$ 3.21 & 81.16 $\pm$ 2.93 \\
    \hline
    \end{tabular}}
}
\caption{Inception Scores ({\bf IS}) and Fr\'{e}chet Inception Distances ({\bf FID}). IS and FID are computed from 20 splits with 1000 images in each split. Unsupervised baselines fail to edit source images to a desired attribute strength and show classification accuracies close to a random guess (around 20\%), however, they have misleadingly high IS and low FID scores (because changes are subtle compared to the source images).}
\label{tab:baseline_metric}
\end{table*}
\section{IS and FID Scores}
Additional Inception Scores (IS)~\cite{salimans2016improved}, Fr\'{e}chet Inception Distances (FID)~\cite{heusel2017gans} are reported in Table~\ref{tab:baseline_metric}.
Classifiers for evaluating classification accuracies are also used to compute Inception Scores and as auxiliary classifiers in training Disc-CGAN/IPCGAN. The unsupervised baselines have high Inception Scores and low Fr\'{e}chet Inception Distances but very low classification accuracies since their outputs are almost identical to source images. Collectively, PC-GAN demonstrates comparable performance with fully-supervised baselines and are significantly better than unsupervised methods.

\section{Network Architectures}
We show the architectures of our Elo ranking network as well as the spatial transformer network in Table~\ref{tab:arch_elo}. Facial attribute classifiers are finetuned ResNet-18~\cite{he2016deep}.
\begin{table}[H]
\centering
\scalebox{0.8}{
\begin{tabular}{lrr}
\hline
Layers & Weights & Activations \\ \hline
Input image & & $224\times 224 \times 3$ \\
\texttt{ResNet-18 features} & & $7\times 7 \times 512$ \\
\texttt{conv}, \texttt{pad} $1$, \texttt{stride} $1$ & $3\times 3 \times 64$ & $7\times 7 \times 64$ \\
\texttt{BatchNorm}, \texttt{LeakyReLU} & & \\
\texttt{conv}, \texttt{pad} $1$, \texttt{stride} $1$ & $3\times 3 \times 1$ & $7\times 7 \times 1$ \\
\texttt{Global AvgPool} & & $1\times 1 \times 1$ \\ \hline
\end{tabular}}
\caption{Architecture of Elo ranking network. \small{\texttt{ResNet-18 features}} are the CNN layers before its classifier.}
\label{tab:arch_elo}
\end{table}

\section{Additional Results}
\begin{figure*}
    \subfloat{%
    \begin{minipage}{\linewidth}
    \centering
    \includegraphics[width=0.44\linewidth]{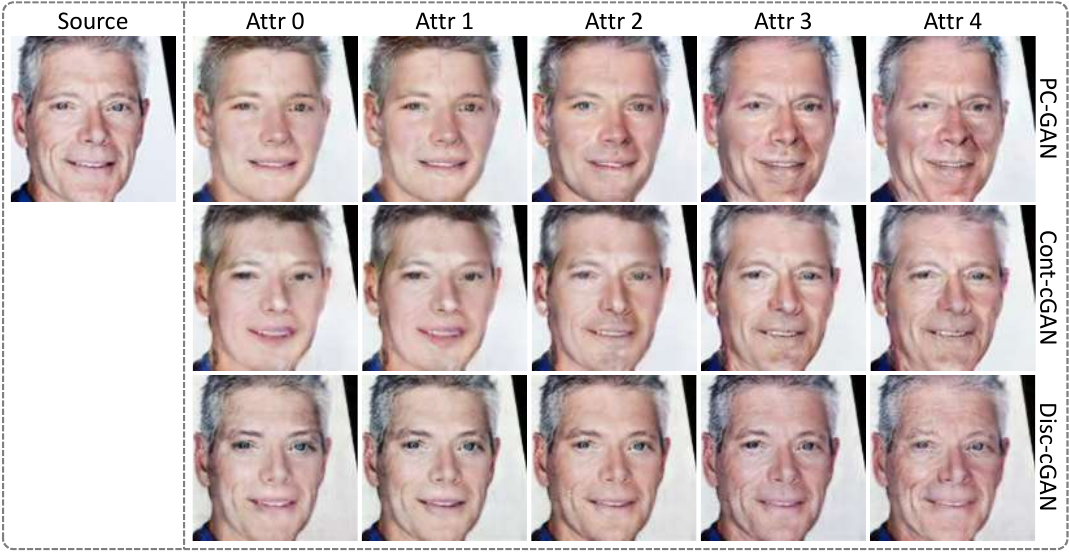}
    \includegraphics[width=0.44\linewidth]{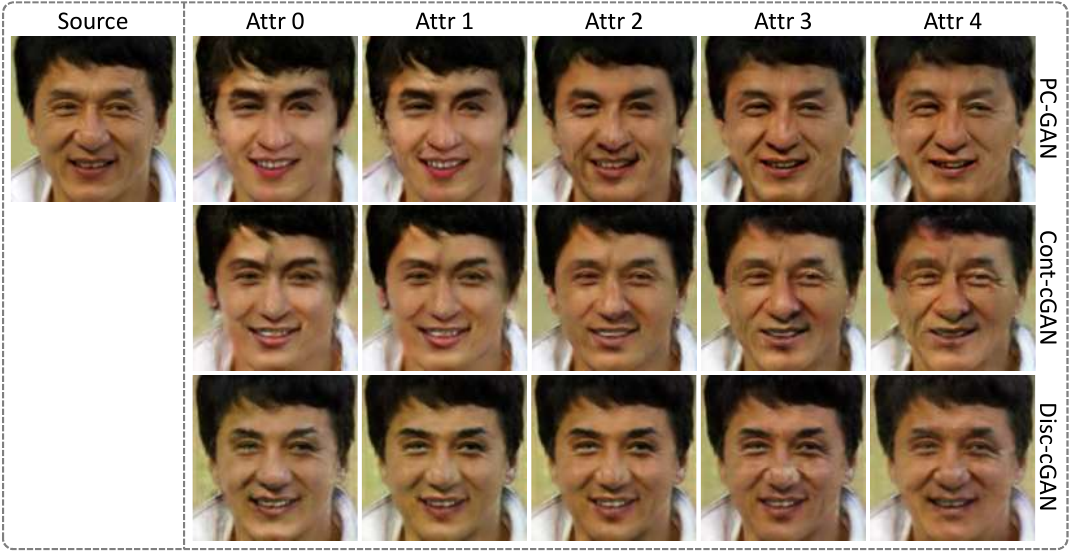}
    \end{minipage}%
    }\par
    \subfloat{%
    \begin{minipage}{\linewidth}
    \centering
    \includegraphics[width=0.44\linewidth]{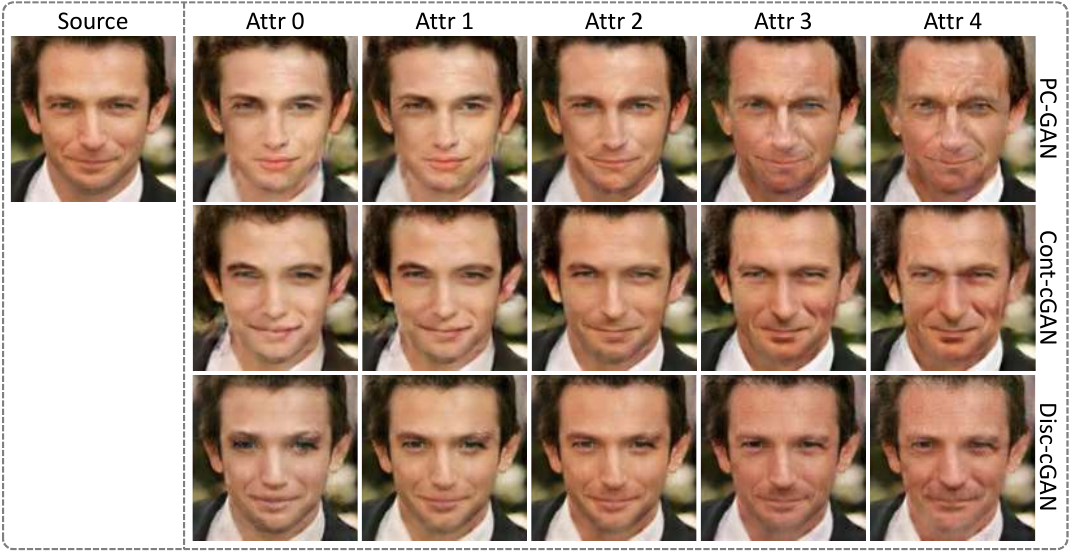}
    \includegraphics[width=0.44\linewidth]{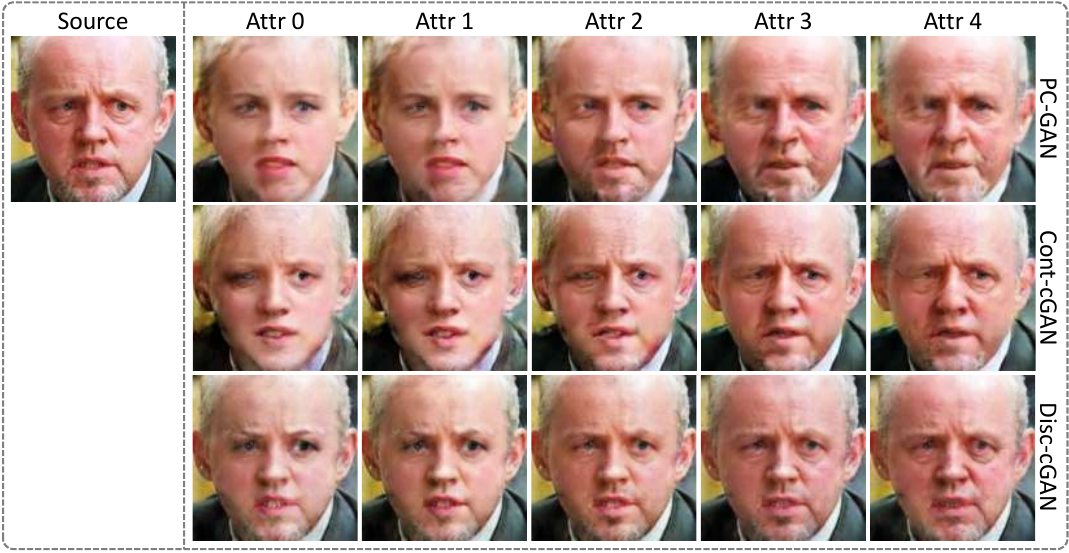}
    \end{minipage}%
    }\par
    \subfloat{%
    \begin{minipage}{\linewidth}
    \centering
    \includegraphics[width=0.44\linewidth]{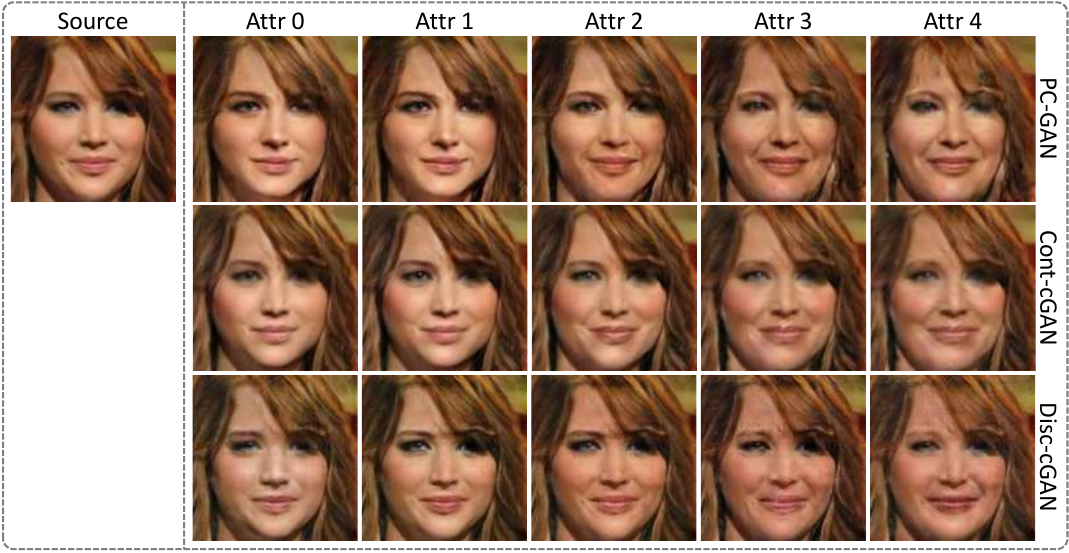}
    \includegraphics[width=0.44\linewidth]{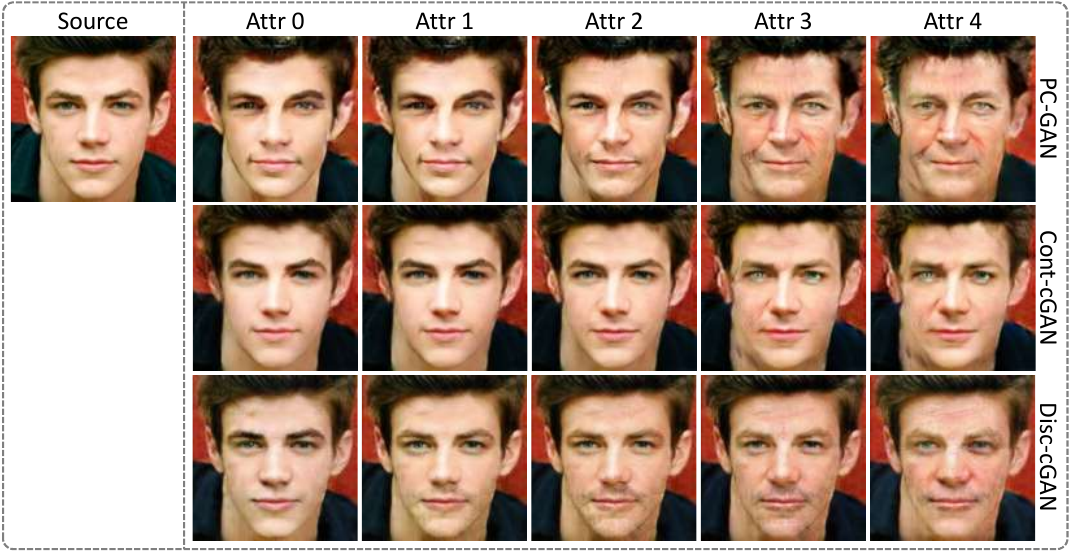}
    \end{minipage}%
    }\par
    \subfloat{%
    \begin{minipage}{\linewidth}
    \centering
    \includegraphics[width=0.44\linewidth]{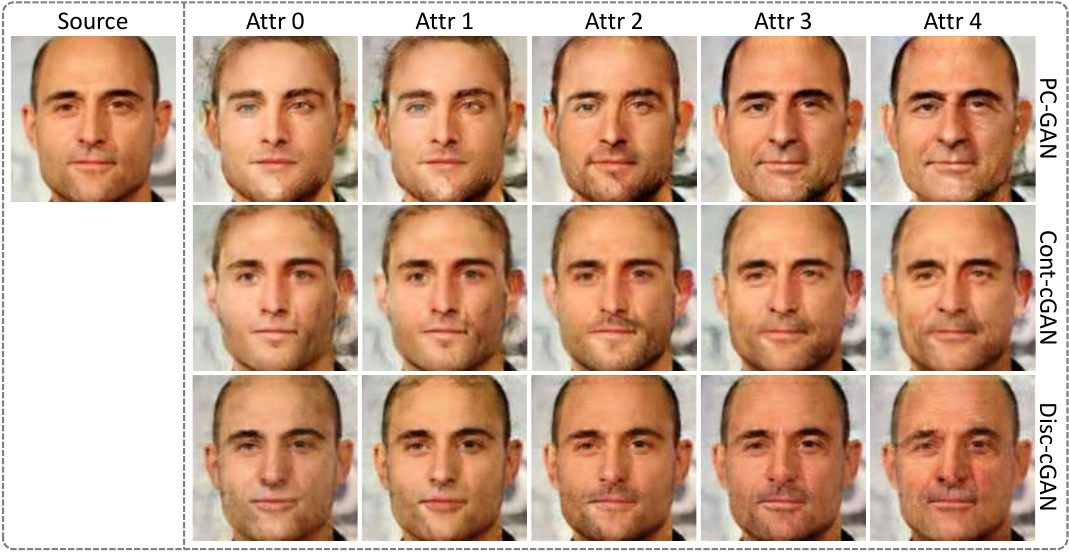}
    \includegraphics[width=0.44\linewidth]{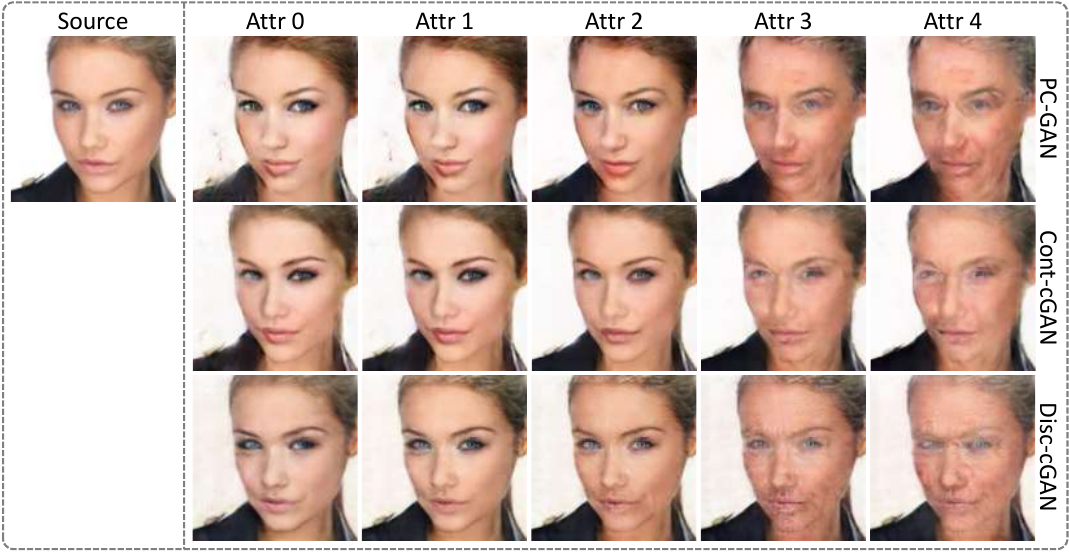}
    \end{minipage}%
    }\par
    \subfloat{%
    \begin{minipage}{\linewidth}
    \centering
    \includegraphics[width=0.44\linewidth]{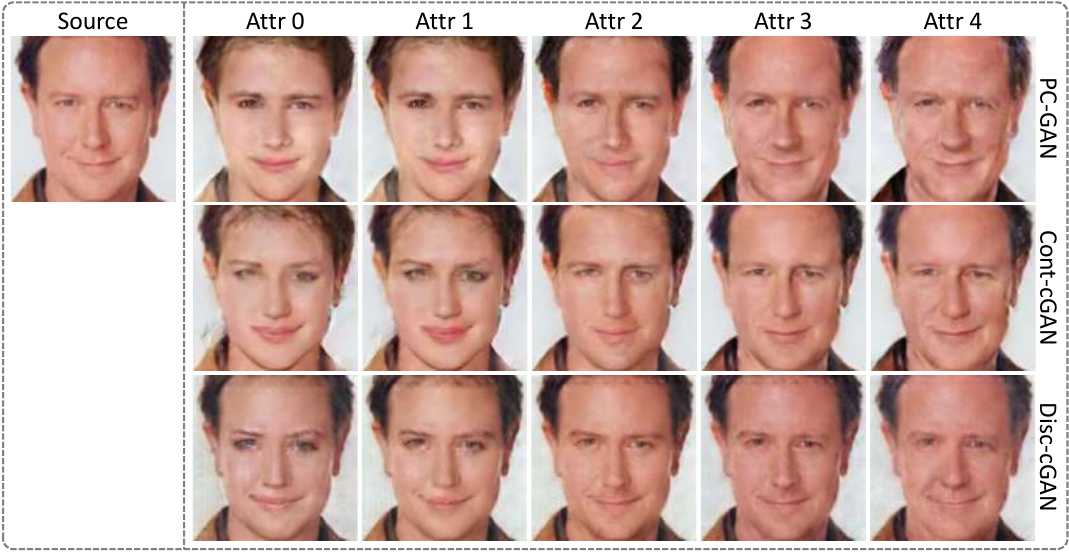}
    \includegraphics[width=0.44\linewidth]{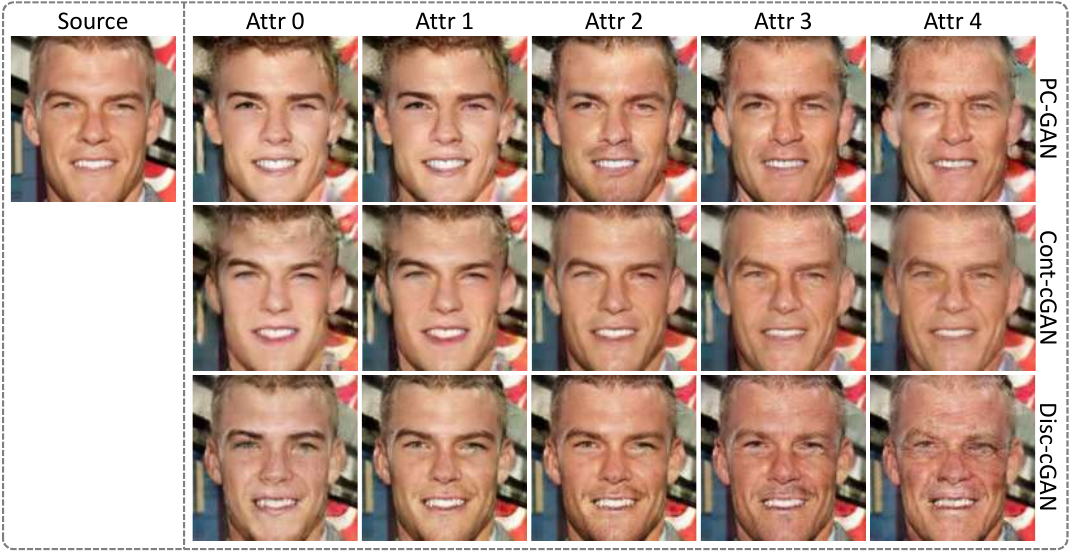}
    \end{minipage}%
    }\par
    \caption{Comparison of PC-GAN with Cont-CGAN and Disc-CGAN on the CACD dataset. Attribute values from \texttt{Attr0} to \texttt{Attr4} correspond to age of $15$, $25$, $35$, $45$ and $55$, respectively.}
    \label{fig:comp_cacd}
\end{figure*}

\begin{figure*}
    \subfloat{%
    \begin{minipage}{\linewidth}
    \centering
    \includegraphics[width=0.44\linewidth]{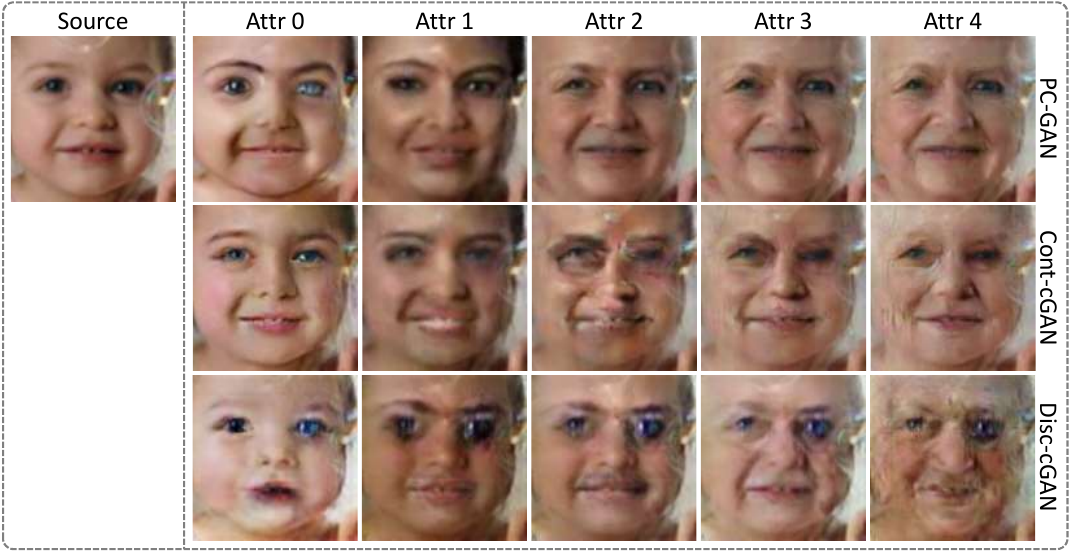}
    \includegraphics[width=0.44\linewidth]{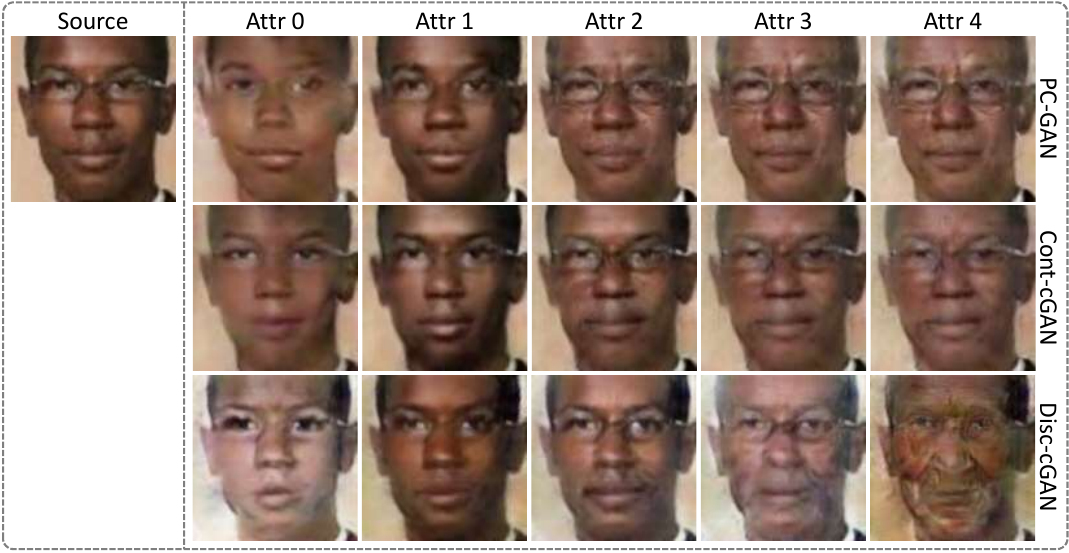}
    \end{minipage}%
    }\par
    \subfloat{%
    \begin{minipage}{\linewidth}
    \centering
    \includegraphics[width=0.44\linewidth]{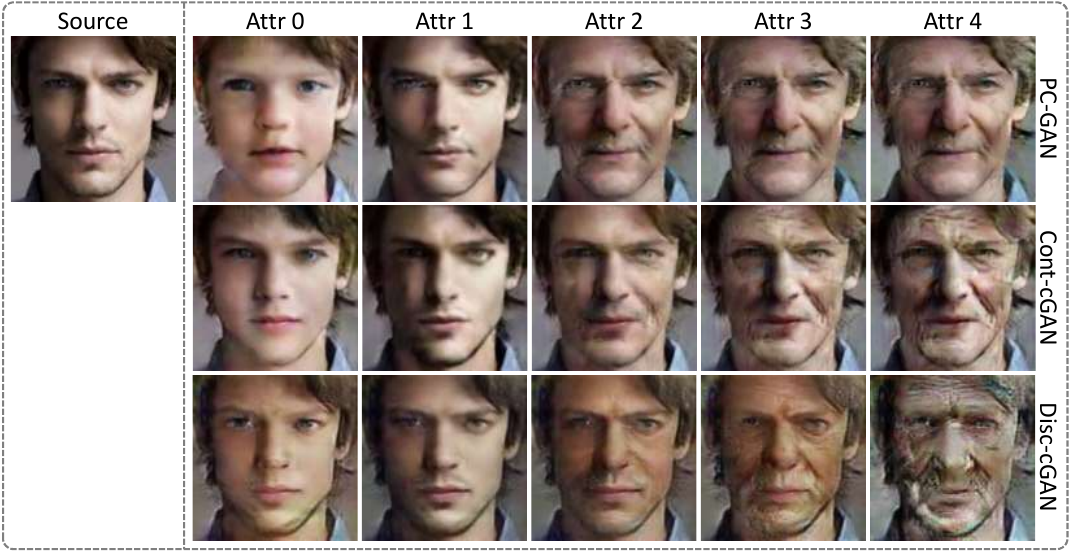}
    \includegraphics[width=0.44\linewidth]{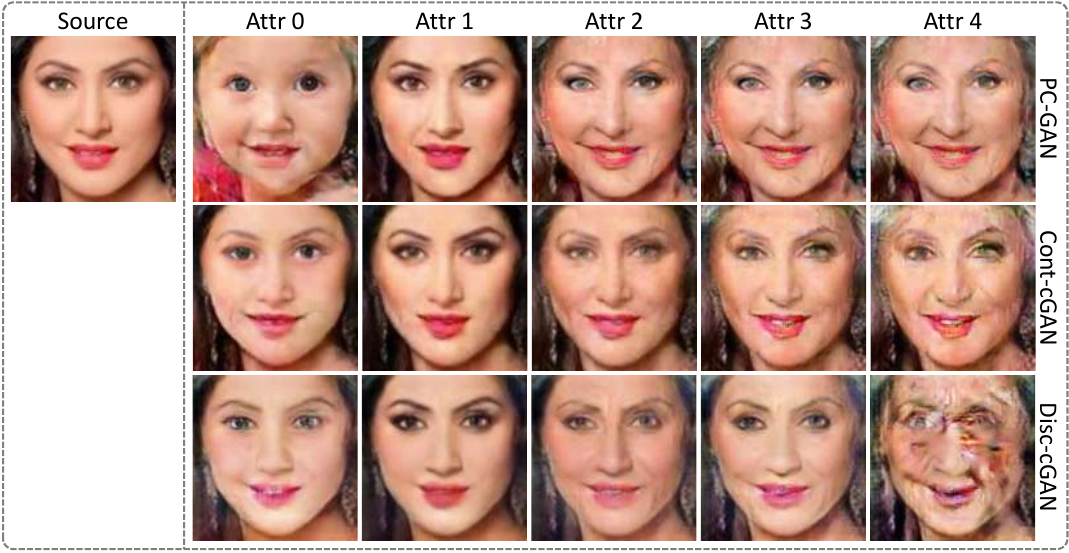}
    \end{minipage}%
    }\par
    \subfloat{%
    \begin{minipage}{\linewidth}
    \centering
    \includegraphics[width=0.44\linewidth]{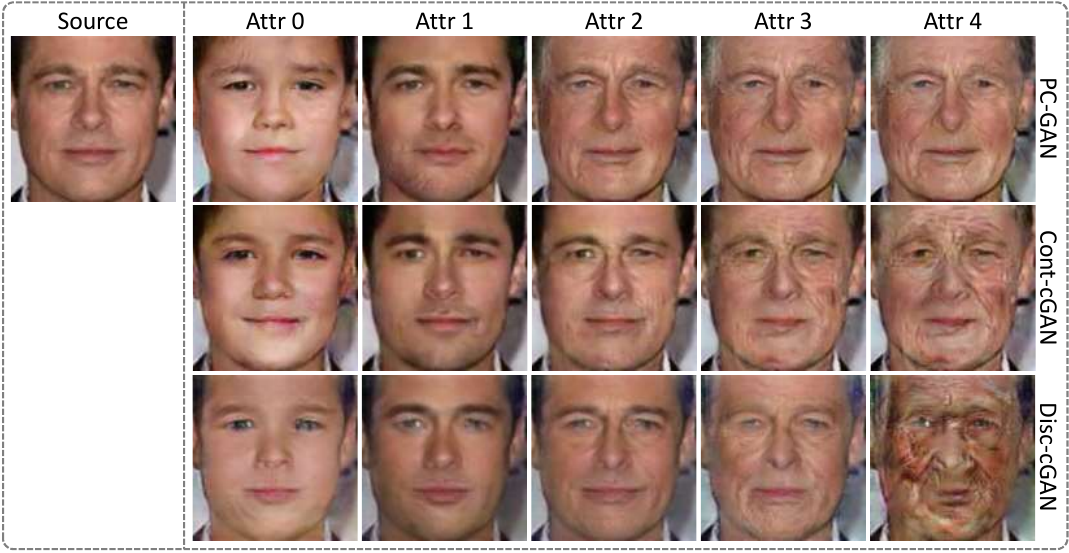}
    \includegraphics[width=0.44\linewidth]{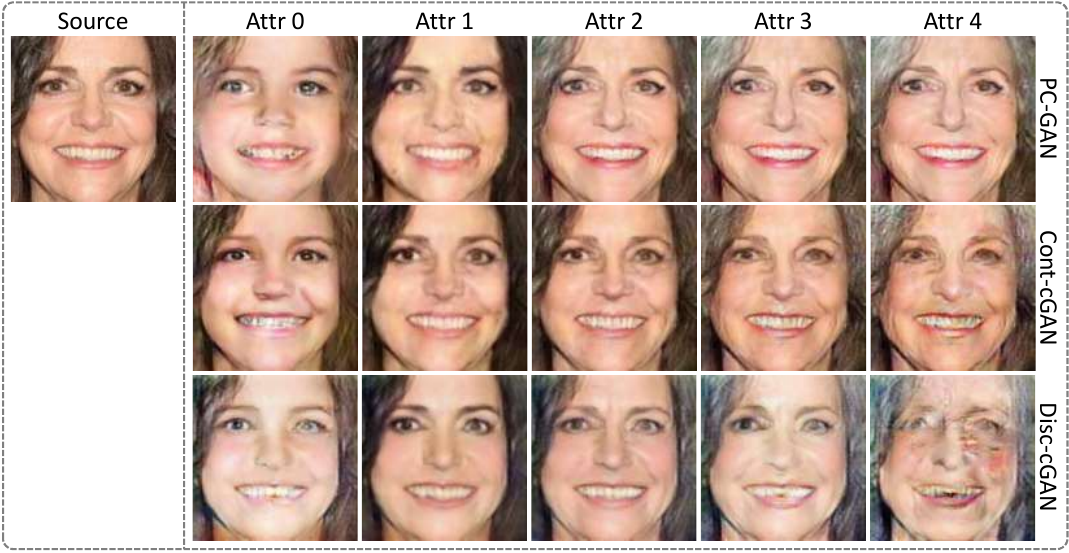}
    \end{minipage}%
    }\par
    \subfloat{%
    \begin{minipage}{\linewidth}
    \centering
    \includegraphics[width=0.44\linewidth]{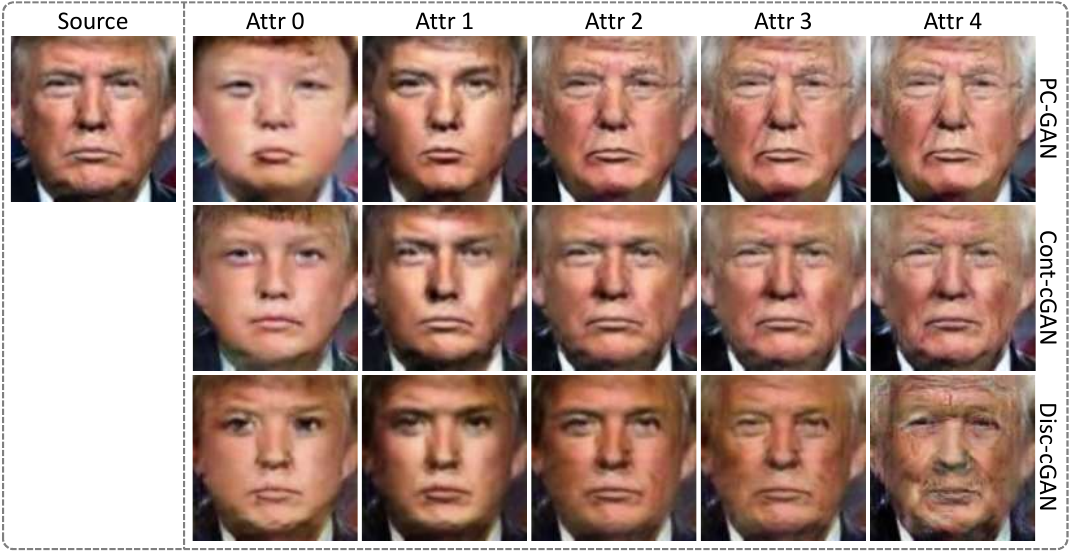}
    \includegraphics[width=0.44\linewidth]{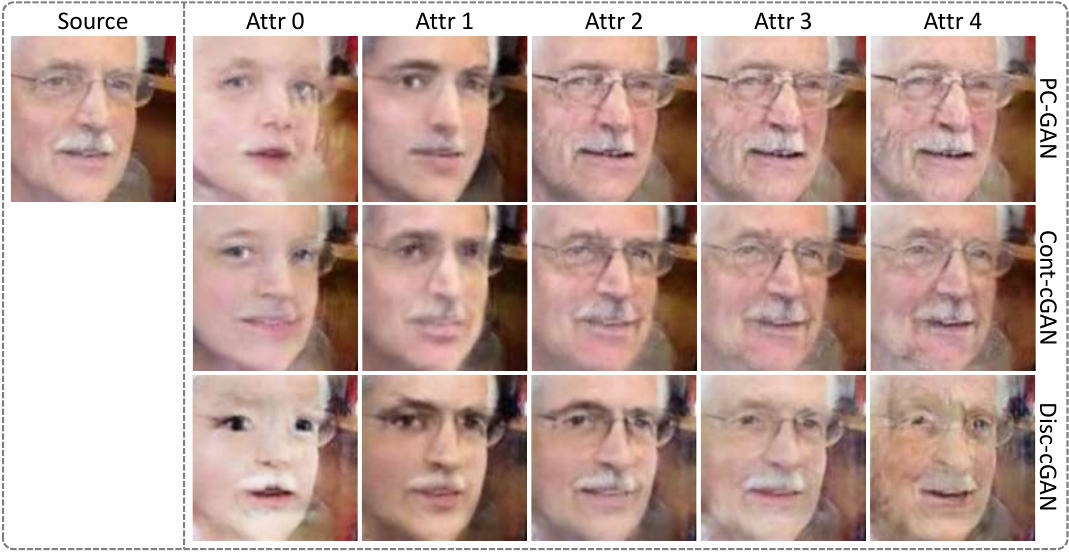}
    \end{minipage}%
    }\par
    \subfloat{%
    \begin{minipage}{\linewidth}
    \centering
    \includegraphics[width=0.44\linewidth]{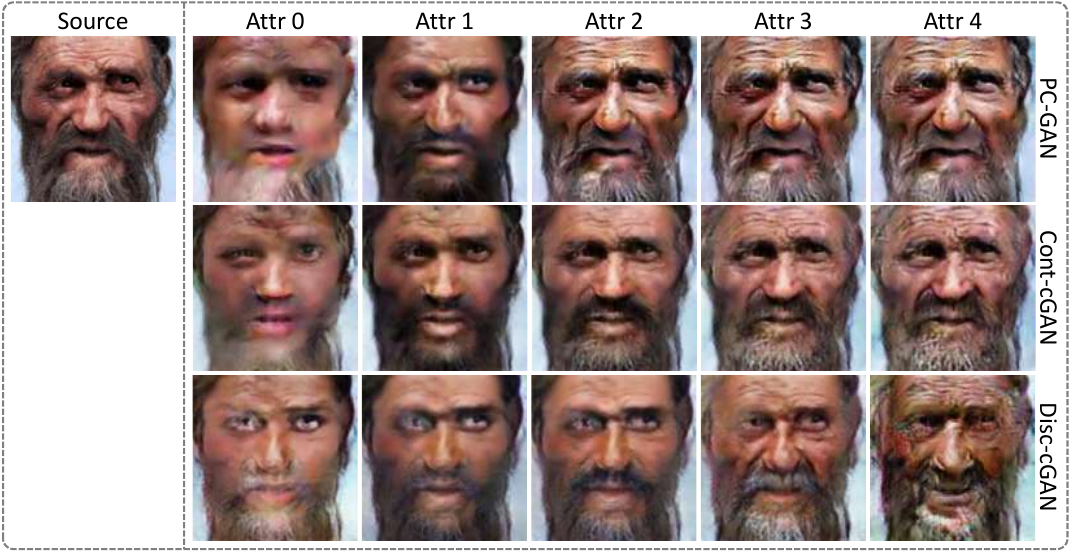}
    \includegraphics[width=0.44\linewidth]{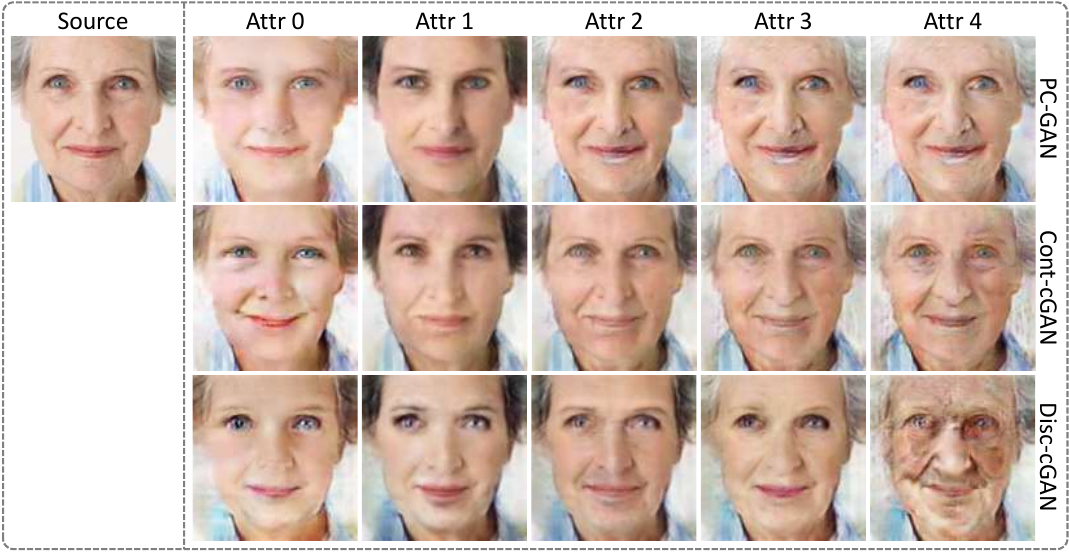}
    \end{minipage}%
    }\par
    \caption{Comparison of PC-GAN with Cont-CGAN and Disc-CGAN on the UTKFace dataset. Attribute values from \texttt{Attr0} to \texttt{Attr4} correspond to age of $10$, $30$, $50$, $70$ and $90$, respectively.}
    \label{fig:comp_utk}
\end{figure*}

\begin{figure*}
    \subfloat{%
    \begin{minipage}{\linewidth}
    \centering
    \includegraphics[width=0.44\linewidth]{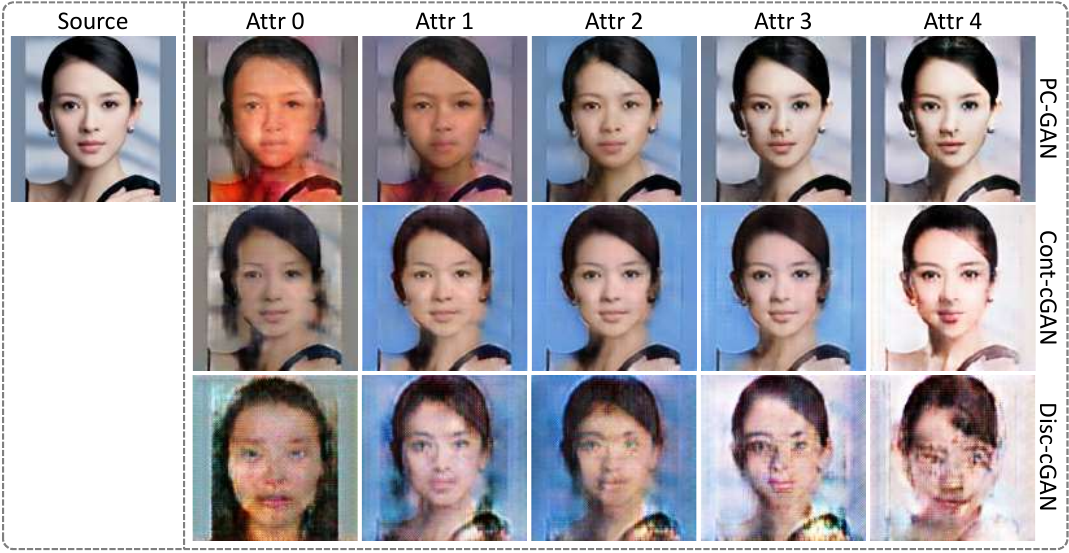}
    \includegraphics[width=0.44\linewidth]{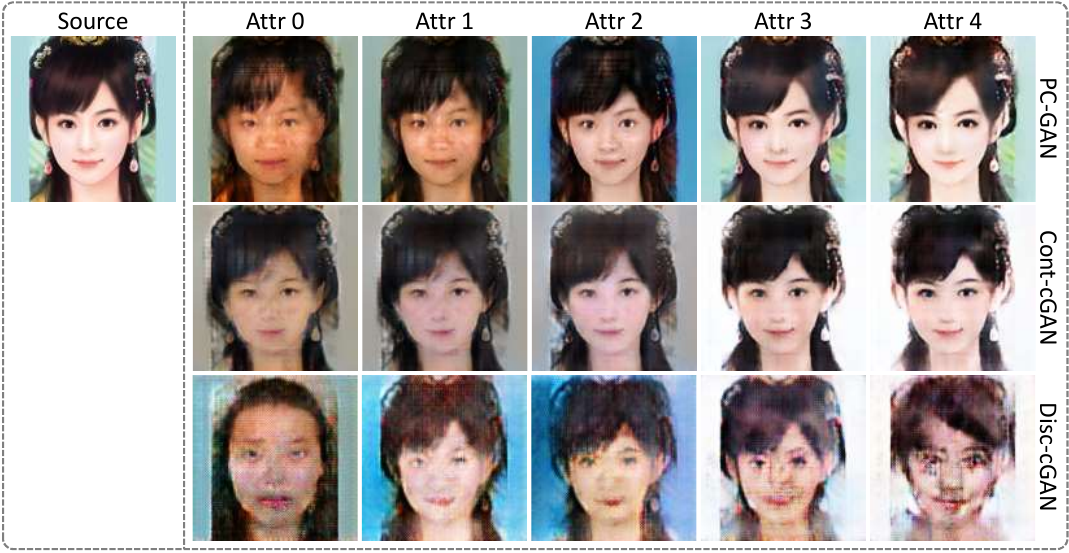}
    \end{minipage}%
    }\par
    \subfloat{%
    \begin{minipage}{\linewidth}
    \centering
    \includegraphics[width=0.44\linewidth]{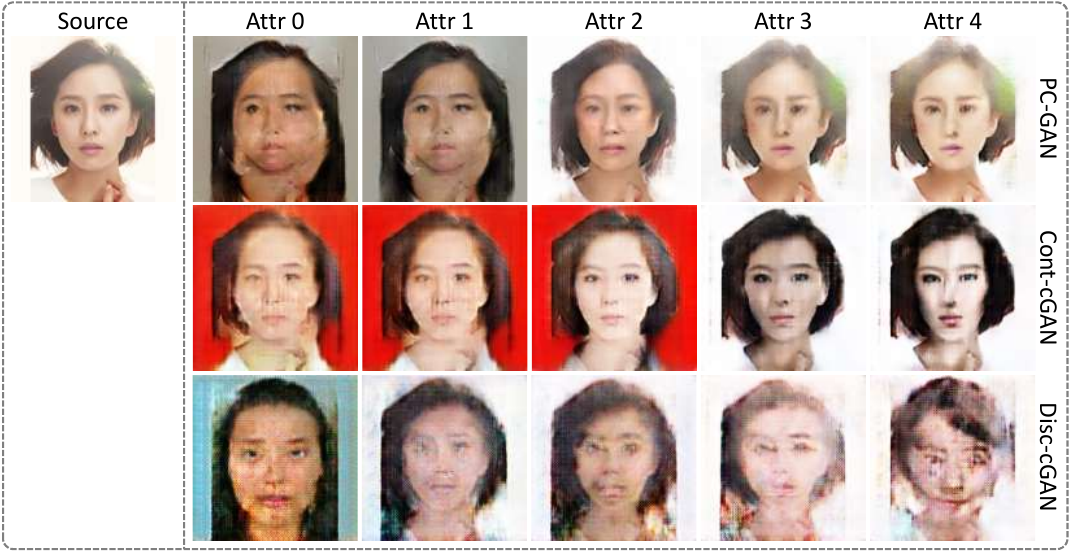}
    \includegraphics[width=0.44\linewidth]{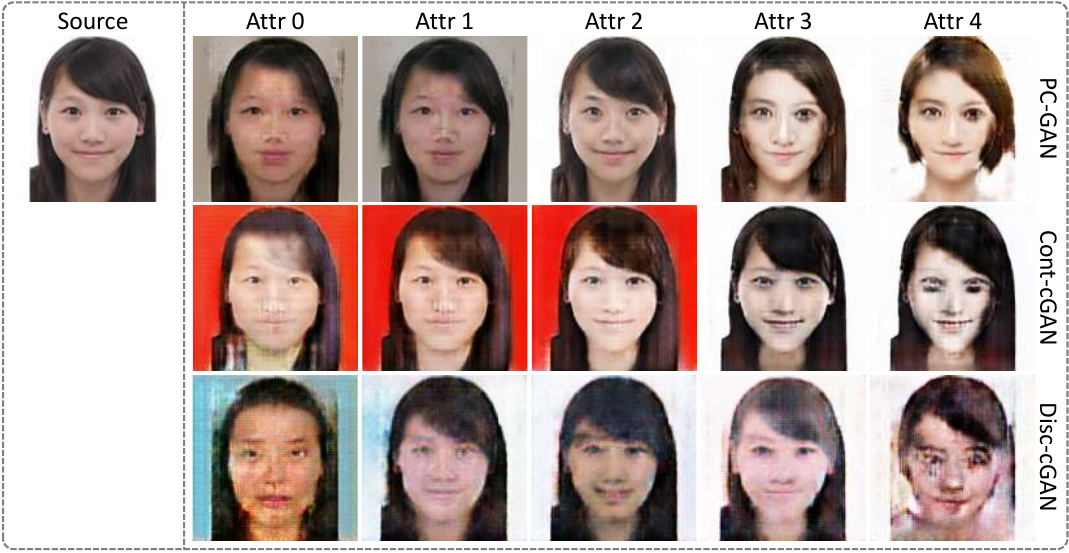}
    \end{minipage}%
    }\par
    \subfloat{%
    \begin{minipage}{\linewidth}
    \centering
    \includegraphics[width=0.44\linewidth]{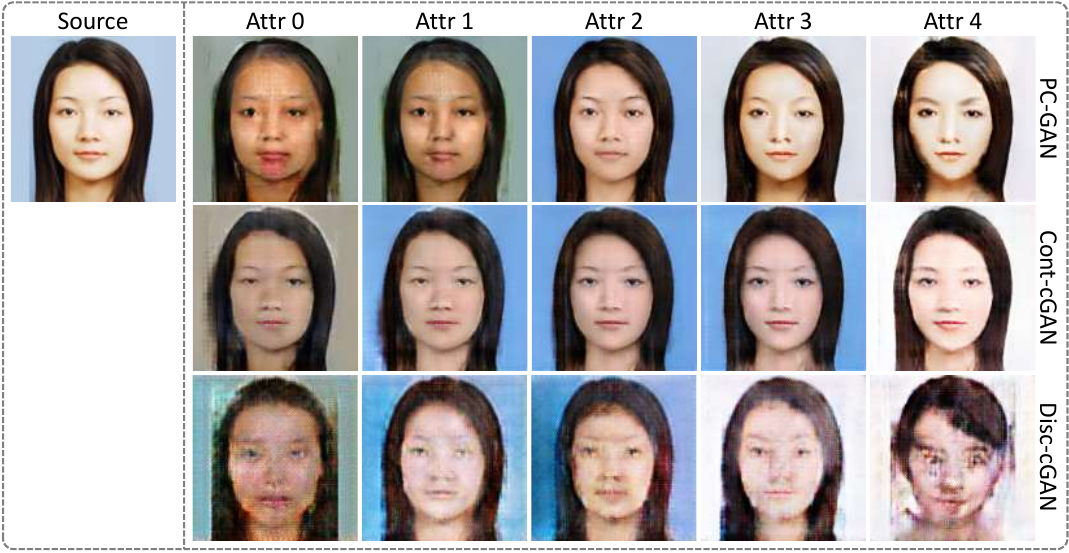}
    \includegraphics[width=0.44\linewidth]{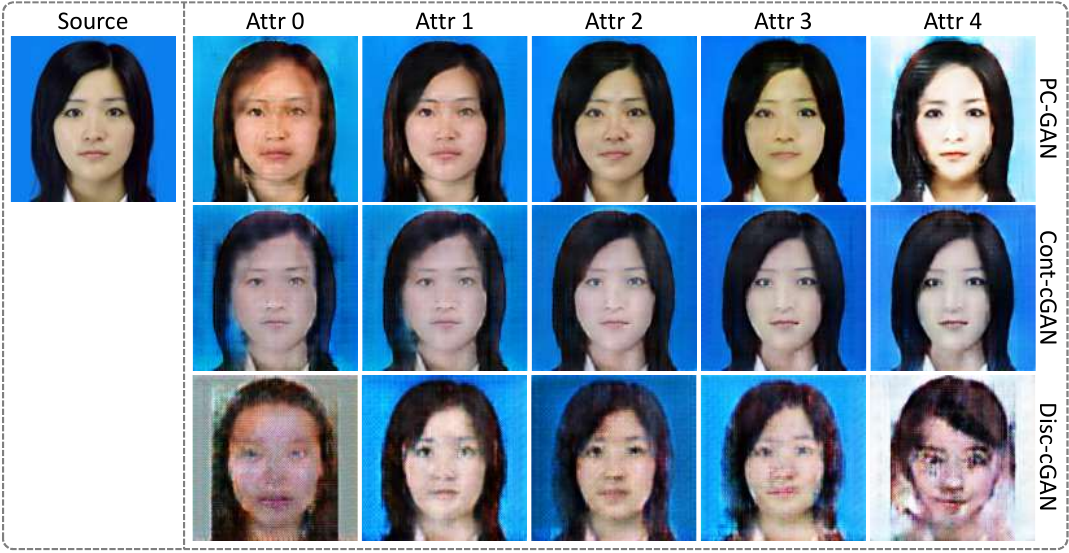}
    \end{minipage}%
    }\par
    \subfloat{%
    \begin{minipage}{\linewidth}
    \centering
    \includegraphics[width=0.44\linewidth]{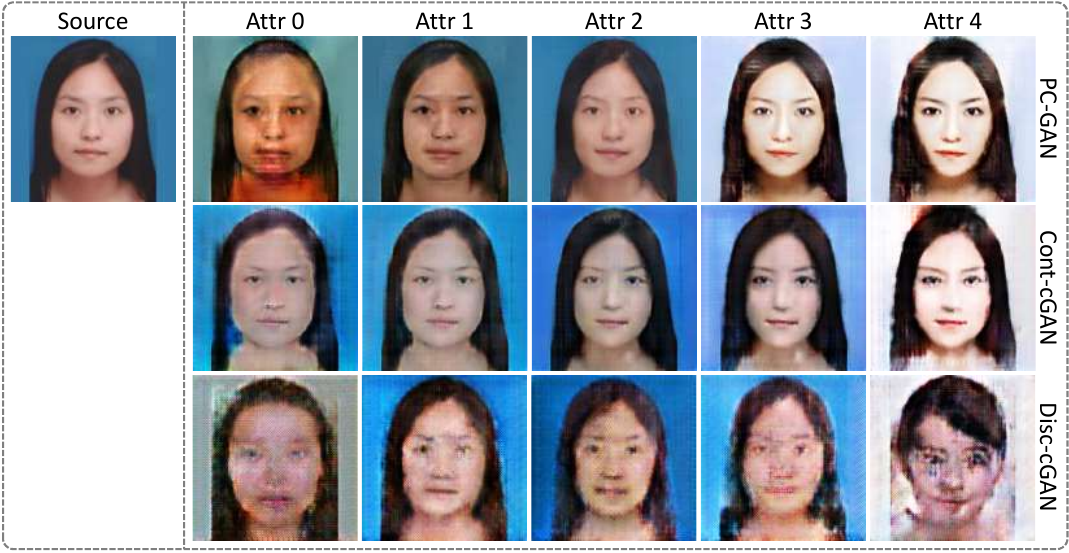}
    \includegraphics[width=0.44\linewidth]{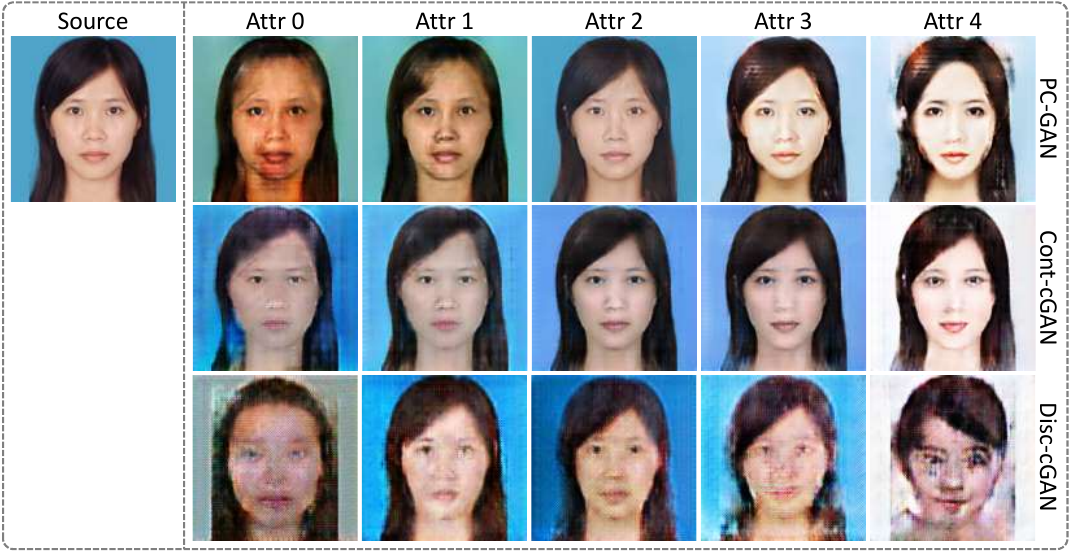}
    \end{minipage}%
    }\par
    \subfloat{%
    \begin{minipage}{\linewidth}
    \centering
    \includegraphics[width=0.44\linewidth]{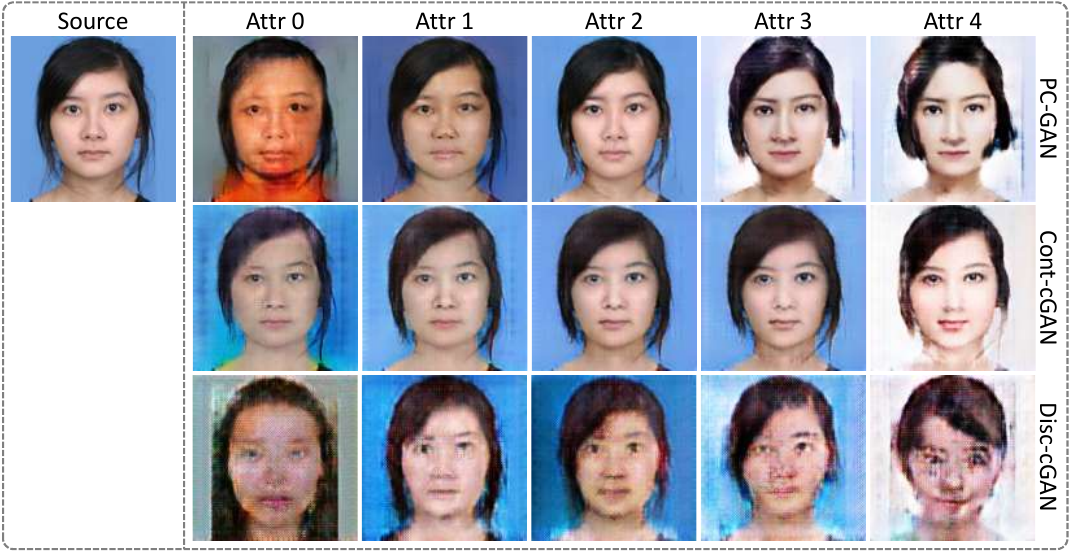}
    \includegraphics[width=0.44\linewidth]{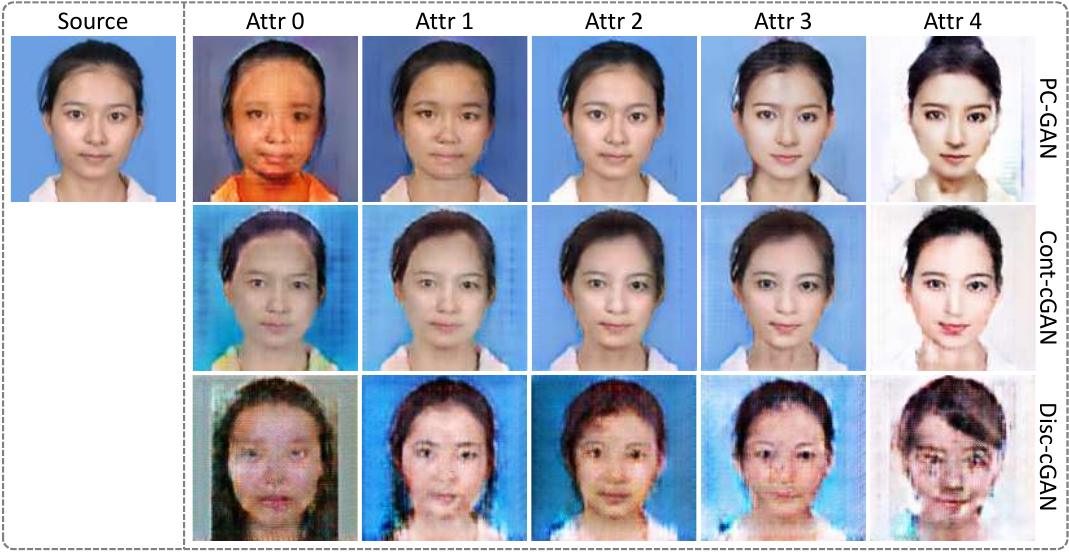}
    \end{minipage}%
    }\par
    \caption{Comparison of PC-GAN with Cont-CGAN and Disc-CGAN on the SCUT-FBP dataset. Attribute values from \texttt{Attr0} to \texttt{Attr4} correspond to score of $1.375$, $2.125$, $2.875$, $3.625$ and $4.5$, respectively.}
    \label{fig:comp_yan}
\end{figure*}

Additional results of our PC-GAN and two fully-supervised baselines Cont-CGAN and Disc-CGAN/IPCGAN~\cite{wang2018face} on CACD, UTKFace, and SCUT-FBP datasets are given in Figure~\ref{fig:comp_cacd}, \ref{fig:comp_utk}, and \ref{fig:comp_yan} respectively. Results for unsupervised baselines are not shown since the changes in outputs are subtle. For CACD, attribute values (from \texttt{Attr0} to \texttt{Attr4}) correspond to ages of $15$, $25$, $35$, $45$ and $55$; for UTK, attribute values correspond to ages of $10$, $30$, $50$, $70$ and $90$; for SCUT-FBP, attribute values correspond to scores of $1.375$, $2.125$, $2.875$, $3.625$ and $4.5$, respectively.

PC-GAN, Cont-CGAN and Disc-CGAN perform similarly on CACD. Disc-GAN performs much worse on UTKFace and SCUT-FBP, presumably due to the discretization of attribute strength. For example, in SCUT-FBP, the number of images are unevenly distributed across discretized attribute groups, that is, groups with least and largest attribute strength (attractiveness) have only limited images. In this case, we are more likely to see mode collapse in Disc-CGAN. As a result, Disc-CGAN is outputting same images for \texttt{Attr0} and \texttt{Attr4} in Figure 8. PC-GAN and Cont-CGAN have a similar quality in synthesized images in all three datasets, which shows PC-GAN can synthesize images of same qualities using pairwise comparisons.

\end{document}